\documentclass[10pt,twocolumn,letterpaper]{article}

\usepackage[pagenumbers]{cvpr}      

\usepackage{graphicx}
\usepackage{amsmath}
\usepackage{amssymb}
\usepackage{booktabs}
\usepackage{pifont}
\usepackage{bbding}
\usepackage{array}
\usepackage{makecell}

\usepackage[pagebackref,breaklinks,colorlinks]{hyperref}
\usepackage{enumitem}
\usepackage{booktabs}
\usepackage{graphicx} 

\usepackage[capitalize]{cleveref}
\crefname{section}{Sec.}{Secs.}
\Crefname{section}{Section}{Sections}
\Crefname{table}{Table}{Tables}
\crefname{table}{Tab.}{Tabs.}

\usepackage{xcolor}

\def\method{\textbf{AS}ymmetri\textbf{C} \textbf{EN}coder \textbf{D}ecoder (\textbf{ASCEND})}

\def\abbre{\textbf{ASCEND}}

\def\cvprPaperID{\textbf{9057}} 
\def\confName{CVPR}
\def\confYear{2023}

\begin{document}

\title{Exploring Vision Transformers as Diffusion Learners}

\author{
    He Cao \textsuperscript{\rm 1,2}\thanks{This work was done when Cao He was intern at IDEA.},
    Jianan Wang \textsuperscript{\rm 1},
    Tianhe Ren \textsuperscript{\rm 1},
    Xianbiao Qi \textsuperscript{\rm 1}, 
    Yihao Chen \textsuperscript{\rm 1}, \\
    Yuan Yao \textsuperscript{\rm 2},
    Lei Zhang \textsuperscript{\rm 1} \\
    \textsuperscript{\rm 1} International Digital Economy Academy (IDEA).\\
    \textsuperscript{\rm 2} The Hong Kong University of Science and Technology. \\
    {\small{hcaoaf@connect.ust.hk}},
    {\small{\{wangjianan, rentianhe, qixianbiao, chenyihao\}@idea.edu.cn}}, \\
    {\small{yuany@ust.hk}},
    {\small{leizhang@idea.edu.cn}}
}

\maketitle

\begin{abstract}
Score-based diffusion models have captured widespread attention and funded fast progress of recent vision generative tasks. In this paper, we focus on diffusion model backbone which has been much neglected before. We systematically explore vision Transformers as diffusion learners for various generative tasks. With our improvements the performance of vanilla ViT-based backbone (IU-ViT) is boosted to be on par with traditional U-Net-based methods. We further provide a hypothesis on the implication of disentangling the generative backbone as an encoder-decoder structure and show proof-of-concept experiments verifying the effectiveness of a stronger encoder for generative tasks with \method. Our improvements achieve competitive results on CIFAR-10, CelebA, LSUN, CUB Bird and large-resolution text-to-image tasks. To the best of our knowledge, we are the first to successfully train a single diffusion model on text-to-image task beyond $64\times64$ resolution. We hope this will motivate people to rethink the modeling choices and the training pipelines for diffusion-based generative models.

\end{abstract}

\begin{figure}[h]
\centering
\includegraphics[width=1.\linewidth]{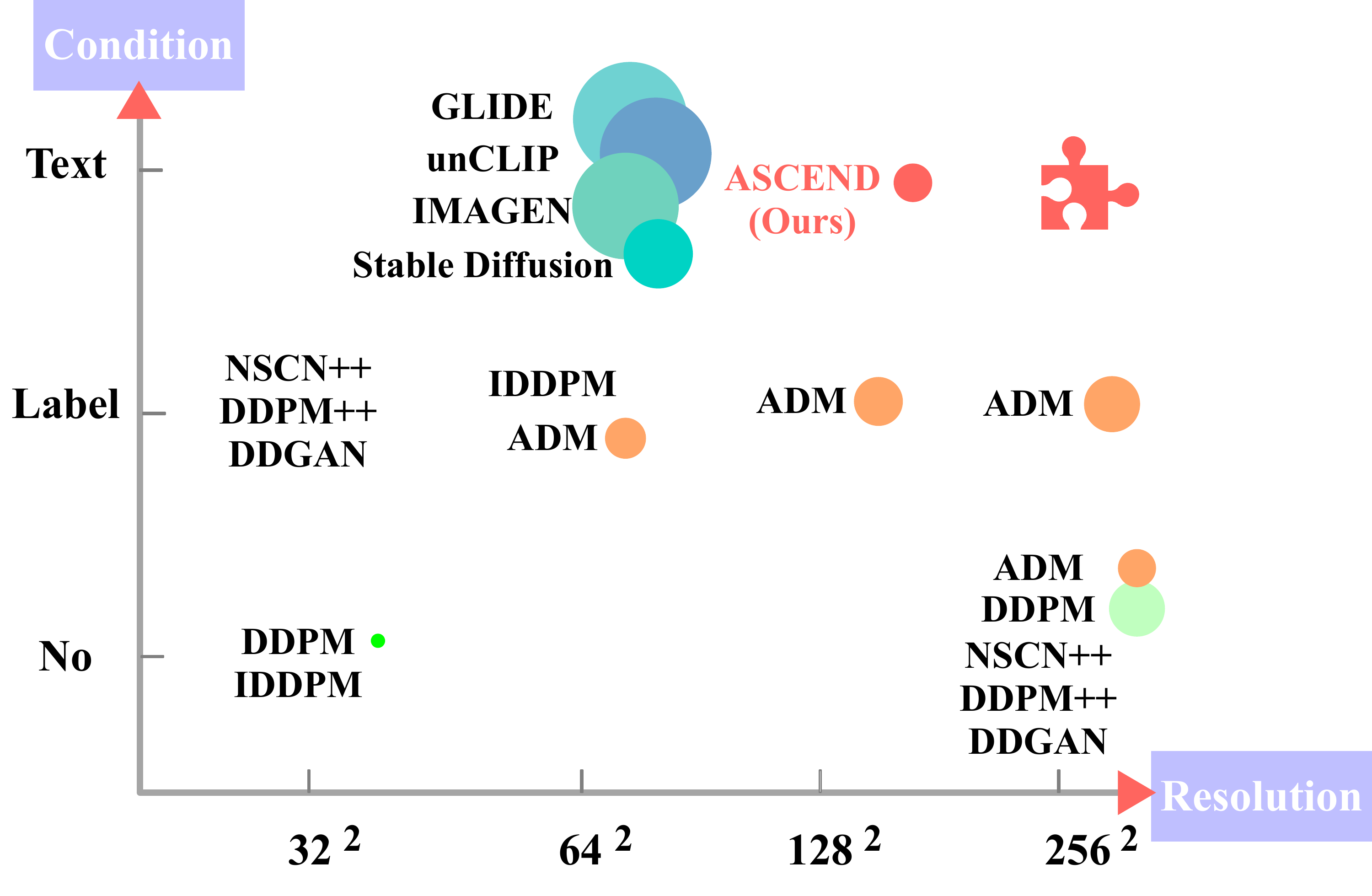}
\caption{Representative diffusion generative models by resolution and conditioning. Red arrow indicates direction of increasing difficulty. We reflect reported model size (excluding conditioning network if any) by the size of colored dots, or otherwise left blank. Our work (ASCEND) is the first to attempt $128 \times 128$ single-stage text-to-image generation while the larger resolution $256 \times 256$ text-to-image generation (marked by puzzle icon) is yet to be explored.}
\vspace{-1em}
\label{fig:generative_tasks}
\end{figure}

\begin{figure*}[h]
\centering
\includegraphics[width=0.85\linewidth]{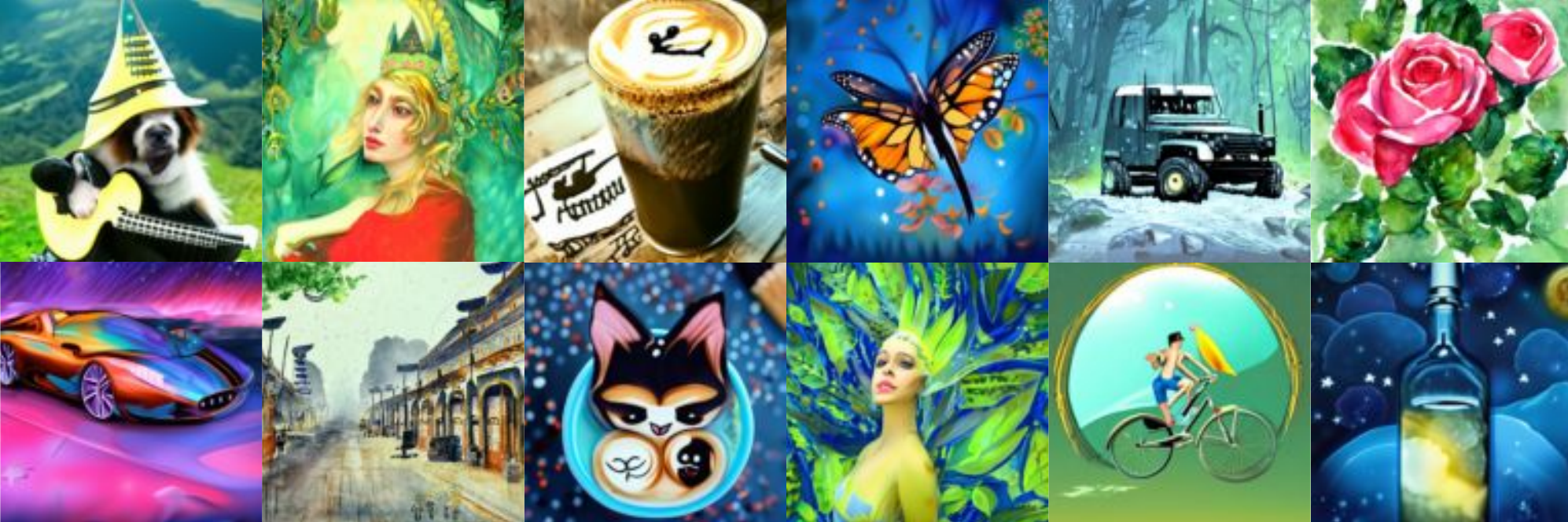}
\caption{$128\times128$ samples generated \emph{without} super-resolution by training a \emph{single-stage} small-sized (590M) text-to-image diffusion model with ASCEND backbone, sampled with 50 steps. Detailed texts and more samples are included
in Appendix C.
}
\label{fig:ascend_t2i}
\end{figure*}

\begin{figure*}[h]
\centering
\includegraphics[width=0.8\linewidth]{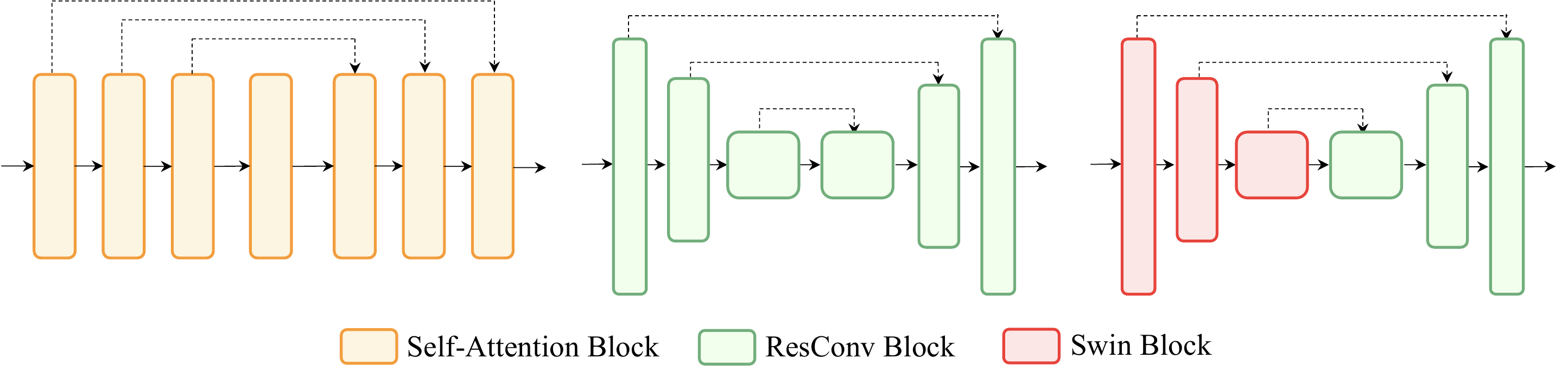}
\caption{Illustrative architecture of U-ViT (Left), U-Net (Middle) and ASCEND (Right). }
\label{fig:encoder_decoder}
\end{figure*}

\section{Introduction}
\label{sec:intro}
Content creation is a hallmark of human intelligence and a long-standing challenge for machine learning algorithms. Recently, text-to-image models such as unCLIP\cite{dalle2}, Imagen\cite{imagen}, and Latent Diffusion\cite{latentdiffusion} have demonstrated an impressive capability of generating photorealistic and creative images given textual instructions. The leap is unprecedented and quickly become viral, attracting widespread public attention. The promise of an era of AI Generated Content (AIGC) is infectious and encourages shared enthusiasm both in the industry and within a larger research community.

The recent progress in image synthesis is enabled by advances in modeling, especially score-based diffusion models. In consequence, the applications and design space of diffusion models have attracted wide research attention. However, the backbone for diffusion models is rarely studied. Moreover, while Transformer has demonstrated great success in both natural language processing and computer vision, there has been little exploration of ViT for generative tasks.

Gen-ViT\cite{genvit} is the first to use standard ViT as diffusion backbone but with poor results. 
Concurrent to our work, DiTs\cite{DiTs} propose a Transformer architecture based on ViT yielding SOTA results on class-conditional ImageNet generation. But DiT learns a simpler distribution of compressed latent space leveraging pre-trained variational autoencoder (VAE \cite{vae}) from Stable Diffusion, hence inherently a multi-stage solution shielding away from increased complexity of high-dimensional distribution learning.
Recently, U-ViT\cite{uvit} shows improved results with long skip connections and a convolutional operation before final prediction, but the experiments focus on small-resolution generation tasks and the performance still lags behind traditional U-Net backbone. 
In our exploration 

we show that with a few improvements on U-ViT (IU-ViT), the performance gap could be further closed.
{To scale to more complex generative problems, we propose \method ~as a scalable one-stage diffusion backbone for larger-resolution unconditional generation and  text-to-image generations.}

{On the training side, pioneering works of text-to-image models quickly set standards for the community: 1) tremendous data and compute are devoted to fuel large generative models with billions of parameters; 2) multi-stage training pipelines are adopted: unCLIP\cite{dalle2} and Imagen\cite{dalle2} generate high-resolution images in a cascaded fashion by first generating low-resolution images followed by separate super-resolution models;} Latent Diffusion \cite{latentdiffusion} first trains a VAE \cite{vae} to compress images to a dense latent space, followed by a second stage to learn the diffusion process in the compressed space. Recent works such as Imagen Video \cite{imagen_video} 
extend image generation to video generation and follow the cascaded pipeline by training 7 models in parallel. Such multi-stage solutions induce a fragile long inference pipeline and are more susceptible to train-test distribution shift. 

{The current design and training paradigm of diffusion models leads to a natural question: \emph{could diffusion models benefit from an end2end training via a better backbone design?}}  More specifically, while U-Net remains the dominant de facto diffusion backbone, vision Transformers have shown great promises in broader vision tasks such as classification \cite{swin,pvt}, detection \cite{detr,dino, dab, dn_detr}, and even low-level segmentation \cite{segformer, rethinking_segmentation}. Compared to CNNs, vision Transformer is generally preferable at large scale because of its scalability and efficiency \cite{vit}. In this paper, we address the former question with ASCEND. ASCEND uses a strong Transformer encoder and a lightweight convolutional decoder. It achieves competitive results on generation tasks such as CIFAR-10, LSUN\cite{lsun}, CelebA\cite{celeba}, CUB Bird\cite{cub-bird} and even the un-attempted task of \textbf{single-stage} \textbf{larger-resolution} \textbf{text-to-image} generation. {Our main contributions can be summarized as follows: }

\begin{itemize}[leftmargin=*, itemsep=2pt]
    \item We reflect on the fast progress of diffusion generative models and propose to systematically measure model capability by orthogonally evaluating the scale of target visual resolution as well as the complexity of external conditioning. 
    \item We thoroughly explore vision Transformers for modeling diffusion scores. We made improvements on U-ViT\cite{uvit} (IU-ViT) which bridges the performance gap between vanilla ViT and U-Net as diffusion backbones for low-resolution generation tasks. Furthermore, we propose to design diffusion backbones with disentangled encoder-decoder architecture and verified that our ASCEND is a scalable diffusion learner. 
    We highlight the strong potential of Transformer-like architectures for unified modeling among vision tasks and encourage the community to explore more data- and compute-efficient training paradigms.
    \item We perform a systematical empirical study on using vision Transformers as diffusion backbones for various generation tasks. Our improved U-ViT (IU-ViT) yields an FID of 2.56 on CIFAR-10 and SOTA FID of 1.57 on CelebA $64 \times 64$. Our proposed hierarchical encoder-decoder model ASCEND is scalable to larger-resolution and multi-modality generation tasks where vanilla ViT-based models struggle for satisfactory results, such as single-stage $128 \times 128$ text-to-image generation.

\end{itemize}

\section{Related Work}
\label{sec:related}

\paragraph{Diffusion Models}
Recently, diffusion models\cite{ddpm,sohl2015deep} have emerged as a promising family of generative models, achieving a state-of-the-art sample quality in various image-generation scenarios. As a class of score-based generative models, diffusion models are inspired by non-equilibrium thermodynamics and contain a forward and a backward process. In the forward process, models gradually add noise to input data according to a predefined schedule, turning data distribution into an isotropic Gaussian. In the reverse process, models learn to invert the noising procedure so that it can turn noise into data at inference. More rigorously, the forward process can be written as adding noise to a clean data $\mathbf{x}_0 \sim p(\mathbf{x}_0)$ in $T$ steps with pre-defined variance schedule $\beta_t$. Each forward transition can be assumed as a Gaussian distribution,
\begin{align}
q\left(\mathbf{x}_{t} \mid \mathbf{x}_{t-1}\right) & = \mathcal{N}\left(\mathbf{x}_{t} ; \sqrt{1-\beta_{t}} \mathbf{x}_{t-1}, \beta_{t} \mathbf{I}\right)
\end{align}
where $\beta_t \in (0,1)$, and the full forward process can be written as,
\vspace{-1.0em}
\begin{align}
q\left(\mathbf{x}_{1: T} \mid \mathbf{x}_{0}\right) & = \prod_{t \geq 1} q\left(\mathbf{x}_{t} \mid \mathbf{x}_{t-1}\right)
\end{align}
The corresponding backward process can be written as,
\begin{eqnarray}
& p_{\theta}\left(\mathbf{x}_{t-1} \mid \mathbf{x}_{t}\right) = \mathcal{N}\left(\mathbf{x}_{t-1} ; \mu_{\theta}\left(\mathbf{x}_{t}, t\right), \Sigma_{\theta}\left(\mathbf{x}_{t}, t\right)\right) \nonumber\\ &= \mathcal{N}\left(\mathbf{x}_{t-1} ; \frac{1}{\sqrt{\alpha_{t}}}\left(\mathbf{x}_{t}-\frac{\beta_{t}}{\sqrt{1-\bar{\alpha}_{t}}} \mathbf{\epsilon} \right),
\frac{1-\bar{\alpha}_{t-1}}{1-\bar{\alpha}_{t}} \beta_{t}\right)
\end{eqnarray}
where $\mathbf{\epsilon} \sim \mathcal{N}(\mathbf{0},\mathbf{I})$, $\alpha_{t}=1-\beta_{t}$, $\bar{\alpha}_{t}=\prod_{i=1}^{t} \alpha_{i}$ and $\theta$ denotes parameters of a neural network learning the denoising objective. The goal of training is to maximize data likelihood $p_{\theta}\left(\mathbf{x}_{0}\right)=\int p_{\theta}\left(\mathbf{x}_{0: T}\right) d \mathbf{x}_{1: T}$, by maximizing the evidence lower bound $\left(\mathrm{ELBO}, \mathcal{L} \leq \log p_{\theta}\left(\mathbf{x}_{0}\right)\right)$. The ELBO can be written as matching the true denoising model $q\left(\mathbf{x}_{t-1} \mid \mathbf{x}_{t}\right)$ with the parameterized $p_{\theta}\left(\mathbf{x}_{t-1} \mid \mathbf{x}_{t}\right)$. During training, given any noised input $\mathbf{x}_t$, the target of the denoising network $\mathbf{\epsilon}_\theta(.)$ is to restore $\mathbf{x}_0$ by predicting the added noise $\epsilon$ via the loss function: 
\begin{align}
\mathcal{L}_{t} & = \mathbb{E}_{\textbf{x}_{0}, \epsilon \sim \mathcal{N}(\mathbf{0},\mathbf{I}) }\left[\left\|\mathbf{\epsilon}-\mathbf{\epsilon}_{\theta}\left(\textbf{x}_0+\sqrt{1-\bar{\alpha}_{t}} \mathbf{\epsilon}, t\right)\right\|^{2}\right]
\end{align}

The applications and design space of diffusion models have attracted wide research attention. 
Many works leverage diffusion model\textquotesingle s gradual process of information discovery for applications such as super-resolution \cite{sr3}, image-to-image translation, \cite{palette} and image editing \cite{sdedit}. Recent explorations on the design space of diffusion models heavily focus on \emph{accelerated sampling}. For example, DDIM \cite{song2021denoising} defines a non-Markovian 
forward process that induces a deterministic generative process with randomness from only the noisiest step, producing high-quality samples 10× to 50× faster than the original formulation. Denoising Diffusion GANs \cite{ddgan} approximate reverse diffusion process by conditional GANs, allowing fast sampling within only a few steps. In \cite{meng2022distillation, progressive}, the authors also explore model distillation to condense the diffusion process. Besides accelerated sampling, people also explore ways of \emph{conditioning} diffusion models for controllable generation, as well as exploring \emph{training} details such as noise scheduling\cite{improved_ddpm} and loss re-weighting\cite{p2loss}.

\paragraph{Standard Diffusion Backbones}
U-Net is by far the go-to choice for parameterizing the denoising network $\epsilon_\theta(\cdot)$. Standard U-Net is an encoder-decoder architecture derived from FCN\cite{FCN}, consisting of compression and expansion paths. The encoder and decoder each operates on the same set of image resolutions with skip connections making each decoder layer aware of features extracted from its corresponding encoder layer. DDPM\cite{ddpm} uses a backbone similar to an unmasked PixelCNN++\cite{pixelcnn_plus_plus} with group normalization \cite{group_norm} throughout. Computation at each spatial size consists of a stack of convolutional residual blocks, downsampling or upsampling blocks, with self-attention blocks applied at pre-specified resolutions.  ADM \cite{beatsgan} explores several architectural changes, such as increasing model depth vs width, using attention at more resolutions, upsampling and downsampling using the BigGAN\cite{biggan} residual blocks and experimenting with adaptive group normalization for incorporating timestep and class information. Imagen\cite{imagen} introduces an Efficient U-Net architecture that shifts model parameters and computation from high-resolution blocks to low-resolution blocks, claiming that the architecture is more memory efficient and induces faster convergence. Other improvements include re-scaling skip connections and reversing the order of downsampling/upsampling operations to increase the speed of forward pass. Even with different variations, the changes to the original U-Net architecture are mild in nature. While U-Net architecture remains dominant in diffusion generative models, some recent works start to explore vision Transformers as an alternative to U-Net, such as Gen-ViT\cite{genvit}, U-ViT\cite{uvit}, DiTs\cite{DiTs}, and Swinv2-Imagen\cite{swinv2imagen}, showing competitive results on small-resolution image generation tasks as well as image super-resolution.

\paragraph{Transformer in Vision and Multi-Modality Tasks} ViT \cite{vit} is the first to demonstrate that a pure Transformer architecture can achieve competitive performance on image classification with large-scale pre-training. DeiT \cite{deit} then proposes a data-efficient training scheme showing that ViT can achieve superior performance compared to modern convolutional neural networks (CNNs). Concurrent works of PVT \cite{pvt}, Swin-T \cite{swin}, and MViT \cite{mvit} reintroduce multi-scale hierarchies into Transformer following the spatial configuration of a typical convolutional architecture such as ResNet-50, making vision Transformers more suitable for dense predictions, such as object detection and semantic segmentation. To further improve the generalization ability on datasets across all scales, more works such as ConViT \cite{convit}, CvT \cite{cvt}, and CoAtNet \cite{coatnet} attempt to incorporate the inductive bias of CNNs into Transformer models via enacting attention within local receptive fields or by extending the FFN layers with implicit or explicit convolutional designs. Recent works in the multi-modal realm \cite{vilt, trar, probing, empirical} further demonstrate the ability of Transformers to model interactions across different modalities by leveraging flexible and high-capacity self-attention or cross-attention computations.

\section{Method}
\label{sec:method}
We start our systematical exploration 
by setting clear standards for assessing the complexity of diffusion generative tasks. We then improve on previous works of vanilla ViT-based diffusion backbone to further close its gap with traditional U-Net-based backbone. Since pioneering works \cite{genvit, uvit} suggest that vanilla ViT architecture could potentially benefit large scale or cross-modality diffusion training, we experimentally verify and analyze whether this hoped-for scalability is warranted. Finally we propose our hierarchical encoder-decoder network as an efficient and scalable diffusion learner.

\subsection{Difficulty Diagram for Vision Diffusion Models}
\label{subsec:generation-tasks}
We observe that by far, diffusion backbone is fairly convoluted with generation tasks: previous works on unconditional or label-conditional generations typically evaluate on self-chosen datasets of different resolutions, while recent works on text-to-image generation usually flash out remarkable sample images out of a generation pipeline which typically involves 2-3 independent models.  We start our exploration of diffusion backbones by first disentangling vision generative tasks by target resolution and conditioning as shown in Figure ~\ref{fig:generative_tasks} with representative diffusion works. Note that we only consider single-stage pure generative models here, image-to-image models and super-resolution models are not included. We hope that clearly defining the difficulty and dimensions of generation tasks would encourage fairer comparisons among diffusion model backbones.

\subsection{Vanilla ViT-based Diffusion Backbone }
\label{explore uvit}
\paragraph{Revisiting Diffusion Backbones} The diffusion model shares great resemblance to stacked denoising auto-encoders, where each diffusion step is analogous to a single denoising auto-encoder: it takes a noisy input and makes it slightly less noisy. Popular backbones used in diffusion models are almost exclusively based on the U-Net architecture. An intuitive reason using U-Net for denoising is to use the encoder for information extraction and compression, filtering out irrelevant noise with information bottleneck, and then reconstruct the purified information with the decoder.  

Our exploration of using Vision Transformers as diffusion learners start with U-ViT \cite{uvit}. 
U-ViT makes a few improvements over the standard ViT \cite{vit} to be more suitable for modeling the diffusion objective. Following the standard ViT models, U-ViT considers everything as tokens, including time embedding, label embedding, and noised image patches. By adding extra long skip connections and a $3\times3$ convolutional block before the final output, U-ViT shows reasonable performance on illustrative small-resolution generative tasks. However, the proposed U-ViT model still lags behind U-Net-based diffusion models. In this work, we demonstrate that the performance gap between SOTA U-Net diffusion models and U-ViT can be further reduced with only a few improvements.

\paragraph{IU-ViT for Better Unconditional Generation}
\label{iuvit}
Here we present our improved U-ViT (named IU-ViT) with better performance on low-resolution unconditional generation tasks. Despite the fact that the self-attention mechanism could effectively model global interactions with large scale data, it inherently lacks a localization mechanism to model fine-grained information within small regions. Such locality is highly crucial for dense visual applications like image synthesis since it is much related to the structures like edges, shapes, objects, etc. To enhance the local modeling capability of U-ViT without incurring excessive additional computations, we simply introduce a depth-wise convolution layer into the feed-forward network (named DWConv-FFN) to bring more locality into the transformer structure.

\begin{figure}[h]

\centering
\includegraphics[width=0.90\linewidth, height=5.5cm]{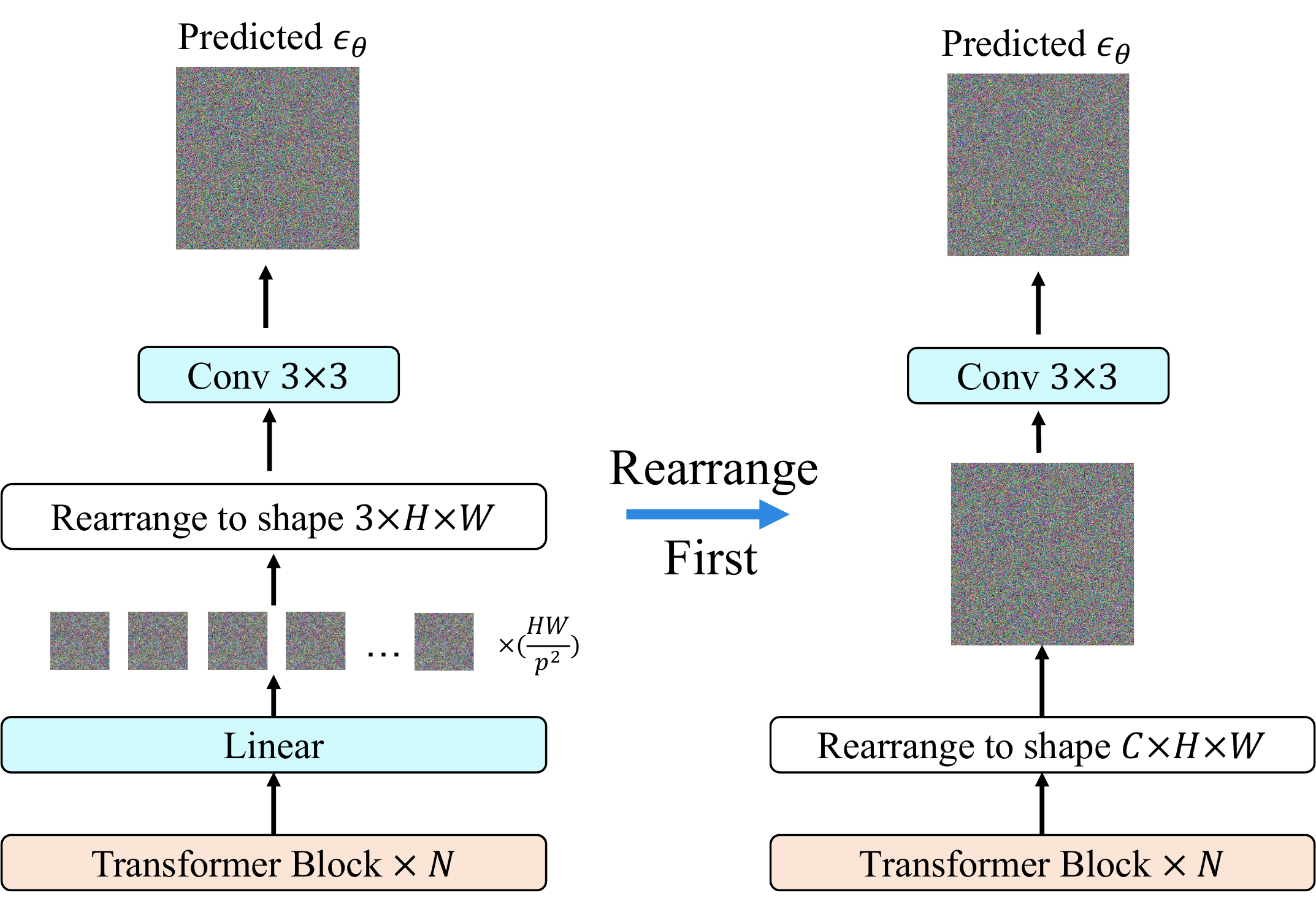}
\caption{Comparison between the Linear-first prediction head (Left) and our Rearrange-first prediction head (Right).}
\label{fig:Improved_Head}
\end{figure}

For the output computation head, U-ViT directly reduces token dimensions with a linear projection and then rearranges the tokens to the shape of input images. Then a $3\times3$ convolution is applied to produce the final predictions. The Linear-first operation used in U-ViT is likely to induce excessive information loss. We propose a Rearrange-first approach instead. Figure ~\ref{fig:Improved_Head} provides a schematic visualization of our design: the output features of the final Transformer block are first rearranged to the shape of $(C, H, W)$ where $C$ denotes token dimension, followed by a $3\times3$ convolution layer to output the final predictions in the shape of $(3, H, W)$. With the introduced improvements IU-ViT achieves better performance on low-resolution unconditional generation.

\paragraph{Is Vanilla ViT-based Model Scalable to more Complex Generative Tasks?}
\label{iuvit_challenges}
After aligning the performance of IU-ViT to U-Net-based models in the regime of low-resolution unconditional generation tasks, we further explore the capacity of our improved U-ViT (IU-ViT) on high-resolution generation and text-to-image tasks as suggested in \cite{genvit,uvit}

For high-resolution generation, since the complexity of self-attention is quadratic to image size, for practicality we increase the patch size for larger-resolution images to ensure the number of input tokens remains constant. 

For text-to-image generation, a typical procedure is to first extract text encodings from a training or pre-trained text encoder. The textual information is then injected into the diffusion backbone via attention operation. Following GLIDE\cite{glide}, the standard approach is to concatenate keys and values of image tokens and text tokens, then perform a fused ``self''-attention. Despite being simple and computationally efficient, this design enforces attention operation to perform both intra-modality and inter-modality information aggregation, making it a more challenging optimization objective. In contrast, we dedicate separate self-attention and cross-attention computations into a transformer block, allowing the self-attention module to focus on interactions of image tokens and the cross-attention module to focus on fusing the token embeddings of different modalities.  

However, even with the aforementioned improvements, we still observe \textit{significant challenges} applying vanilla ViT-based model such as IU-ViT to large-resolution image generation and text-to-image tasks. For large-resolution image generation, we observe that the generated images share an obvious patch effect with little structural information in the early training stage, requiring a very long training time to alleviate the patch effect (as shown in Figure \ref{fig:trainstepcomparison}). Even after many iterative steps, the generated images still lack fine details compared to standard U-Net architecture\cite{ddpm}. For text-to-image generation (64 $\times$ 64 resolution), IU-ViT is able to learn abstract concepts and styles but without fine details and compositional structure, evidently inferior to baseline U-Net.  

We experimented with larger models via increasing transformer depth or token feature dimension, but the above mentioned problems are still persistent. We believe that a backbone architecture based on vanilla ViT may not be a practical solution when scaling up to more complex generative modeling tasks. In summary, IU-ViT shares similar limitations with vanilla ViT\cite{vit}:
\begin{itemize}[leftmargin=*,itemsep=2pt]
    \item Following the vanilla ViT design, IU-ViT \textit{lacks convolutional inductive bias}. Although stacked global self-attention is highly expressive and flexible, some works\cite{swin, NesT} have suggested that ViT can only surpass the performance of CNNs by being pre-trained on large scale data. From the perspective of efficiency and practicality, IU-ViT in limited data regime is able to learn global interactions (featured by abstract concept and style) but struggles when modeling fine-grained instance or regional-level information to compose a high-quality generation.

    \item IU-ViT-like backbone \textit{lacks hierarchical structure} and consequently \textit{without} explicit multi-level hierarchical representations. ViT models maintain a full-length token sequence across all layers, prone to over-smooth representation learning and feature redundancy \cite{DBLP:journals/corr/abs-2104-12753}. Even with reintroduced skip connections, it is still not as efficient as within a hierarchical network.
    In a hierarchical U-Net-like architecture, each decoder layer combines high-level information as well as encoded representations from its corresponding encoder layer with the same spatial size, capable of attending to high-level semantics as well as fine details, while ViT lacks a clear definition of hierarchical correspondence.
\end{itemize}

\subsection{ASCEND: Towards a Single-stage High-resolution Diffusion Model}

\label{subsec:enc-dec}
In this subsection, we explore pushing the frontier of diffusion generative modelling towards end2end single-stage training via a more efficient and higher-capacity backbone design.

\paragraph{Diffusion Backbone with an Encoder-Decoder Perspective:}

Both IU-ViT and U-Net can be broadly viewed as symmetric encoder-decoder architecture with skip connections. The encoder progressively extracts features from inputs and the decoder makes dense predictions to recover injected noises or equivalently denoised inputs given extracted features. According to our previous analyses in \ref{iuvit_challenges}, we consider \textbf{Hierarchical-Encoder-Decoder} architecture with dense skip connections to be desirable diffusion backbone considerations for image generation tasks.

\paragraph{Asymmetric Encoder Decoder}
The de facto diffusion backbone is U-Net which mainly builds upon convolutional blocks, especially ResBlocks \cite{he2016deep}. Convolutional blocks are highly efficient and capable of extracting low-level features, but are less competent at modeling global semantics and structure. In contrast, Transformer with self-attention is highly flexible at capturing long-range relationships, making ViT a tempting backbone choice. However, we observe significant challenges in extending vanilla ViT to more complex vision generative tasks. Firstly, the diffusion model requires the employed backbone to make dense predictions at pixel level. Visual elements can vary substantially in scale, deeming ViT-like fixed scale modeling without hierarchical feature maps unsuitable. Secondly, it is intractable for ViT to model high-resolution images, as the computational complexity of its self-attention is quadratic to image size.

To address the aforementioned limitations of convolutional networks and vanilla ViT, we resort to combining the best of Transformer and traditional CNNs: we rely on the remarkable modeling capacity of Transformer for building hierarchical feature maps and leverage the local bias of traditional CNNs for recovering information from pyramidal feature maps. As a result, we develop an asymmetric encoder-decoder architecture similar to MAE \cite{he2022masked}. We use a strong encoder with Swin Transformer blocks \cite{swin}, which is good at modeling both high-level semantic information and low-level image details, and use a lightweight convolutional decoder to predict dense diffusion objectives from latent representations, as shown in Figure \ref{fig:encoder_decoder}. We call this
\textbf{AS}ymmetri\textbf{C} \textbf{EN}coder \textbf{D}ecoder (\textbf{ASCEND}), and argue that the encoder and decoder architecture can be flexibly combined in a manner that is independent of each other.

\section{Experiments}
We experiment on using vision Transformers as diffusion backbones following the categorization of generative task difficulty according to Section \ref{subsec:generation-tasks}. We first evaluate our improved U-ViT (IU-ViT) on CIFAR-10 and CelebA $64\times64$ showing that the introduced improvements are effective for low-resolution generation tasks. We then verify if such vanilla ViT-based model could 
scale to larger-resolution and cross-modality training as suggested in \cite{genvit, uvit}. Finally we show that our proposed hierarchical \abbre ~network is a more scalable backbone choice for diffusion models with evaluations on $256\times256$ high-resolution generation and $128\times128$ text-to-image generation.

For evaluation, we rely on the standard Fréchet Inception Distance (FID) score for low-resolution generation. But it is widely known that FID as a lump quality evaluation is not ideal for diagnostic purposes \cite{fid, evaluations}. So when exploring larger-resolution generations, we show samples generated during the training process as an intuitive indication of model training behavior with different backbones.

\subsection{Experiments with Improved U-ViT (IU-ViT)}

\paragraph{Improved Low-resolution Unconditional Generation} 
We first evaluate IU-ViT on CIFAR-10. For this 32$\times$32 resolution task, we use a 13-layer ViT of size 45M (on par with U-ViT\cite{uvit}) with two simple improvements as detailed in \ref{iuvit}.

\begin{figure}[h]
\vspace{-1.0em}
\centering
\includegraphics[width=0.9\linewidth, height=5.4cm]{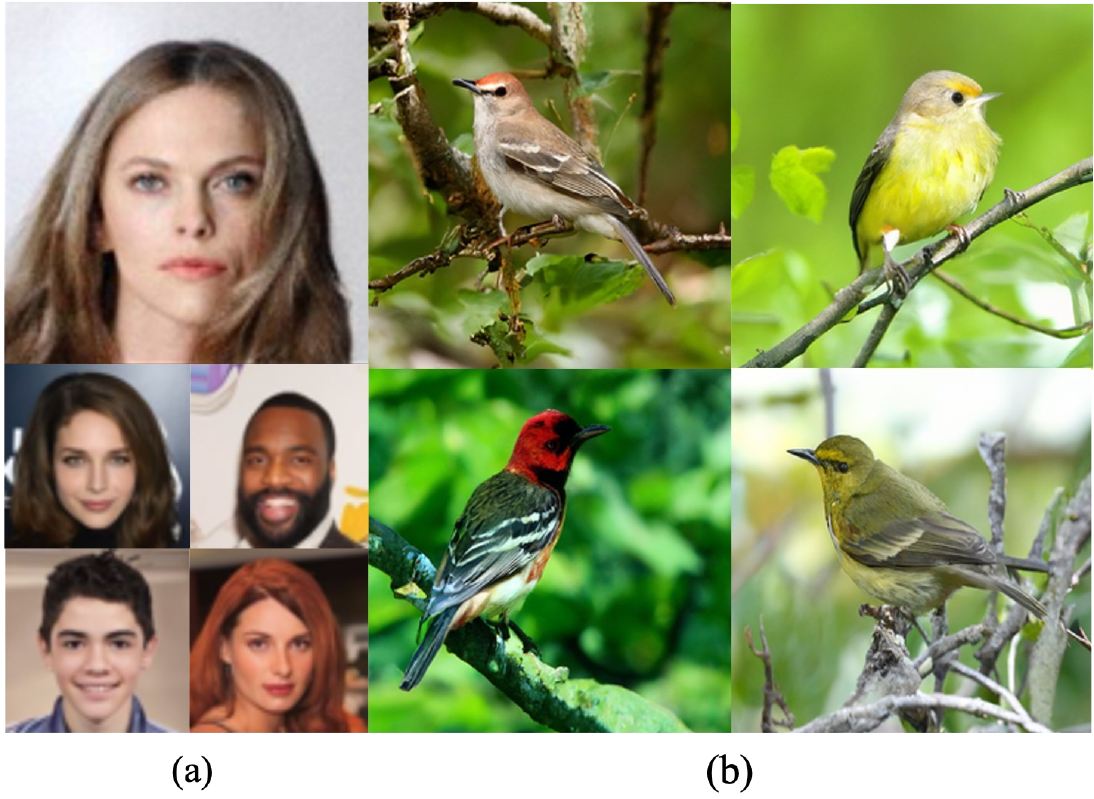}
\caption{Generated samples with (a) IU-ViT on CelebA $64\times64$ and $128\times128$, (b) ASCEND on CUB Bird $256\times256$.}
\label{fig:iuvit_celeba}
\end{figure}

Table \ref{cifar10-generation-results} reports FID scores of competitive models on CIFAR-10, the performance of our Improved U-ViT (IU-ViT) is superior to U-ViT and comparable with U-Net-based diffusion models. And the generation results on CelebA $64\times64$ is reported in Table \ref{celeba-generation-results} which shows that our IU-ViT reaches new SOTA results with 1.57 FID score compared with both U-Net based and U-ViT based models. As shown in Table \ref{uvit-cifar10-ablation}, both Rearrange-first and DWConv-FFN bring positive performance improvements. We also train IU-ViT on CelebA $128\times128$. See Figure \ref{fig:iuvit_celeba} for generated samples.

\begin{figure*}[h]
    \centering
    \includegraphics[width=0.95\linewidth, height=6.5cm]{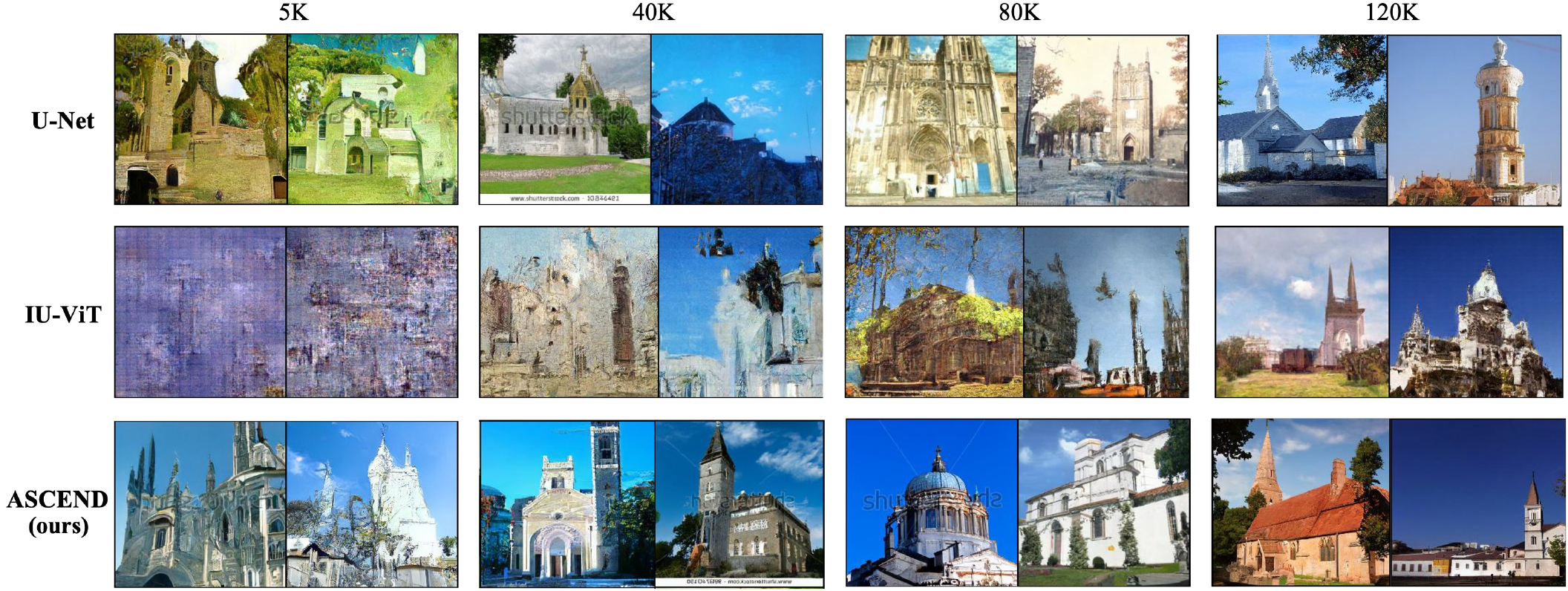}
    
    \caption{Illustrative sample quality during diffusion training with U-Net (top), IU-VIT (middle) and ASCEND (bottom). (1) IU-ViT induces obvious patch effect at early training stage (5K iterations) and lacks fine details even at later training stage. (2) ASCEND is faster at learning structural information as well as fine details than UNet and IU-ViT.}
\label{fig:trainstepcomparison}
\vspace{-1.0em}
\end{figure*}

\begin{table}[]
\centering
\begin{tabular}{l|c|c|c}
\toprule[1.2pt]
Model & FID $\downarrow$ & NFE $\downarrow$ & Params  \\ \hline
TransGAN \cite{transgan}  & 9.26  & 1  & - \\
ViTGAN \cite{vitgan}      & 4.57  & 1  & -\\
StyleGAN2 w/ ADA \cite{stylegan2} & \textbf{2.92}  &  1 &  -\\ \hline
NCSN U-Net \cite{ncsn} & 25.3 & 1000 & -\\
NCSNv2     \cite{ncsnv2} & 10.87 & 1160 & -\\
DDPM U-Net \cite{ddpm}                          & 3.17         & 1000       &   -\\
IDDPM U-Net \cite{improved_ddpm}                & 2.90         & 4000       &   -\\
DDPM++ U-Net \cite{song2021scorebased}             &\textbf{2.55}         & 2000       &      - \\ 
Denoising Diffuison \cite{ddgan}                & 3.17         & 4          &  -\\ \hline
GenViT   \cite{genvit}           & 20.20        & 1000       & 11.6M  \\
U-ViT  \cite{uvit}               & 3.11         & 1000       & 44M  \\ 
Improved U-ViT (ours) & \textbf{2.56}        & 1000        &  45M \\ 
\bottomrule[1.2pt]
\end{tabular}
\caption{Results on CIFAR-10 unconditional generation and model size comparison among ViT-based models. }
\label{cifar10-generation-results}
\end{table}

\begin{table}[]
\centering
\begin{tabular}{l|c|c}
\toprule[1.2pt]
Model    & FID $\downarrow$ & Params  \\ \hline
DDIM (U-Net) \cite{song2021denoising}  & 3.26  & 79M  \\
$\text{Soft Truncation}^{\dag}$  (U-Net) \cite{kim2022soft}      & 1.90  & 62M \\ \hline
U-ViT  \cite{uvit}               & 2.87           & 44M  \\ 
Improved U-ViT (ours) & \textbf{1.57}      &  45M \\ 
\bottomrule[1.2pt]
\end{tabular}
\caption{Results on CelebA 64x64 unconditional generation.}
\label{celeba-generation-results}
\end{table}

\begin{table}
\centering
\renewcommand\arraystretch{1.2} 
\resizebox{\linewidth}{!}{
\begin{tabular}{cc|c} 
\toprule[1.2pt]
 Rearrange-first in Head & use DWConv-FFN & FID  \\ 
\hline 
 \small{\XSolidBrush} &   \small{\XSolidBrush}     &   3.11   \\
    \Checkmark &  \small{\XSolidBrush}      &  2.94    \\
       \small{\XSolidBrush}        &  \Checkmark      &  2.60    \\
     \Checkmark         &   \Checkmark     &  \textbf{2.56}    \\
\toprule[1.2pt]
\end{tabular}
}
\caption{Investigation of our improvements upon U-ViT evaluated on CIFAR-10.}
\label{uvit-cifar10-ablation}
\vspace{-1.5em}
\end{table}

\paragraph{Challenging Scalability to High Resolution} 
To challenge whether vanilla ViT-based model like IU-ViT can synthesize realistic images at larger resolutions, we experimented on generation tasks at resolutions $128\times128$ and $256\times256$. To control computation complexity of attention, we keep the number of image tokens constant (256) across all resolutions by adjusting patch size. For $128\times128$ we use a model of size 442M and for $256\times256$ we use a model of size 527M. See Appendix B for more details.

We observe that as image resolution increases, larger patch size is likely to become a bottleneck for high-quality image generation . For CelebA $128\times128$ with $patch\_size=8$, IU-ViT shows no obvious patch effect. However, when training on $256\times256$ resolution with $patch\_size=16$, it is obvious from Figure ~\ref{fig:trainstepcomparison} that IU-ViT shows noticeable patch effect compared to baseline U-Net at early stage of training iterations (5k), slowly becoming smoother (40k and 80k), but even at later stage of training (120k) it still fails to generate fine details and compositional structures compared to U-Net. We experimented on increased model size via more transformer layers or larger token dimension, but the problem is still persistent. We believe that practically, it is challenging to scale vanilla ViT-based backbone for higher-resolution generation tasks.

\paragraph{Challenging Scalability to Multi-modality}
\label{t2ifailure}

We orthogonally explore whether vanilla ViT-based model is capable of image generation with more complex text conditioning at $64\times64$ resolution.
 
\begin{figure}[h]
\centering

\includegraphics[width=1.0\linewidth]{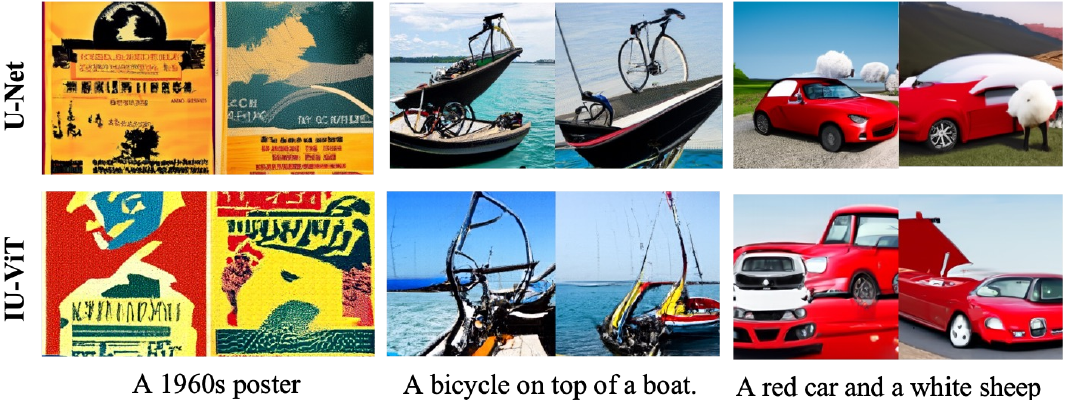}

\caption{$64\times64$ image samples of U-Net (top) and IU-ViT (bottom) on DrawBench\cite{imagen} prompts.}
\label{fig:t2i_UNet_IU-ViT}
\vspace{-0.5em}

\end{figure}

Similarly to Imagen, we use a frozen large-scale pre-trained text encoder (T5-XXL\cite{raffel2020t5}, 4.6B encoder parameters) for text encoding. Unlike Imagen, we do not use fused ``self''-attention  but separate self-attention and cross-attention as introduced in \ref{explore uvit}. In our experiments, we use an IU-ViT network of $\sim$300M trained on Conceptual 12M\cite{cc12m} and compare with an Imagen text-to-image model of similar model size. In Figure \ref{fig:t2i_UNet_IU-ViT}, we show that IU-ViT struggles at learning compositional structure and object-level details, strictly inferior to U-Net at 
text-image semantic alignment.

\subsection{ASCEND: An Efficient Hierarchical Encoder-Decoder Backbone}

\paragraph{Model Settings} In Section \ref{subsec:enc-dec}, we propose that hierarchical encoder-decoder architecture is a viable choice for efficient and practical diffusion learning. Following that principle, we are flexible to adopt a sufficiently powerful feature extraction model as encoder and a less-aggressive model as decoder to reconstruct semantic and fine-grained information at different resolutions. In this section, we first verify this hypothesis on low-resolution generation tasks and then proceed to larger-resolution and cross-modality generative tasks to validate the effectiveness and scalability of such an asymmetric backbone design. 

For implementation, we incorporate SwinTransformer\cite{swin} Block into the encoder and use residual downsampling/upsampling \cite{biggan} to replace patch merging and patch expanding.

For the decoder, we use convolutional operations to gradually make dense predictions utilizing high-level information as well as encoded features from the corresponding encoder layer. We refer to this asymmetric network as \method.

\paragraph{Results and Analyses} 

\begin{itemize}[leftmargin=*,itemsep=2pt]\item \textbf{ASCEND Ablation on CIFAR-10} We conduct extensive ablation study on CIFAR-10 as shown in Table \ref{SECOND-cifar10-ablation}: 1) patch-merging/expanding are less effective than residual down/up-sampling; 2) reducing the number of skip connections impairs model performance; and 3) using SwinBlocks in encoder only is a competent modelling choice.
\begin{table}
\centering
\renewcommand\arraystretch{1.2} 
\resizebox{\linewidth}{!}{
\begin{tabular}{l|c|c} 
\hline
\textbf{Implementation} & \textbf{FID} $\downarrow$ & $\Delta$ \\ 
\hline 
\textbf{ASCEND Baseline} & $\mathbf{2.98}$  & - \\
PatchMerging $\searrow$ and PatchExpanding $\nearrow$ & 12.81 & +9.83 \\
\text{Reduce skip connections} $\downarrow$ & 6.52 & +3.54\\
Swin Encoder + Swin Decoder  & 4.62 & +1.64 \\
Conv Encoder + Swin Decoder & 5.87 & +2.89\\
\hline
\end{tabular}
}
\caption{Ablation results of ASCEND (Swin Encoder + Conv Decoder) on CIFAR-10. 
PatchMerging $\searrow$ \cite{swin} and PatchExpanding $\nearrow$ \cite{swinv2imagen} are for down(up) sampling. }
\label{SECOND-cifar10-ablation}
\vspace{-1.5em}
\end{table}

\item \textbf{Scalability to High-resolution Image Generation} We qualitatively inspect images generated by training diffusion models with three different backbones: U-Net, IU-ViT and ASCEND at different training iterations in Figure \ref{fig:trainstepcomparison}. ASCEND is faster at generating high-quality images than IU-ViT and U-Net, and is noticeably better than IU-ViT at later training stage. We observe similar results on CUB-Bird $256\times256$, see Appendix C for more details. Note that we only train for 120K steps for \textbf{efficiency}.

\item \textbf{Scalability to Text-to-Image Generation Beyond $\mathbf{64}\times\mathbf{64}$} We challenge ASCEND for the un-attempted territory of single-stage high-resolution text-to-image generation. For computation efficiency, we experiment on $128 \times 128$ text-to-image generation with a relatively small-sized model of $\sim$590M parameters (in contrast to standard text-to-image models such as unCLIP and Imagen that use $>2$B  parameters for modeling at $64 \times 64$ as shown in Figure ~\ref{fig:generative_tasks}). Similarly to \ref{t2ifailure}, we use a pre-trained T5-XXL encoder for text encoding and train on LAION-Aesthetics dataset \cite{laion5b}  containing $\sim$120M web-crawled text-image pairs. We show that ASCEND is able to generate high-quality samples as shown Figure~\ref{fig:ascend_t2i} despite of the moderate model size. We hope this will motivate the community to take advantage of developments in vision Transformers to explore more capable encoders for diffusion learning, and to challenge status-quo training paradigms for \textbf{end2end high-resolution text-to-image generation}. 

\end{itemize}

\section{Conclusion and Discussions}
The revolution of backbones plays a central role in advancing vision model capabilities and training paradigms. In this work, we propose to set clear standards for evaluating the capability of diffusion backbones by target resolution and conditioning. We systematically explored vision Transformer architectures as diffusion learners. We 
made improvements on previous work of U-ViT named IU-ViT and demonstrated competitive performance compared to well-tuned U-Net diffusion backbones on CIFAR-10 and CelebA. Noticing the challenges of extending vanilla ViT-based backbone to larger-resolution and multi-modality training, we proposed \method ~as a scalable diffusion learner and showed proof-of-concept performance on high-resolution generation. We also pushed further for unexplored end2end higher-resolution text-to-image generation with encouraging results. We hope this will motivate the community to explore more capable backbones and new training paradigms for more robust and efficient vision generative modeling.

{\small
\bibliographystyle{ieee_fullname}
\bibliography{egbib}

\begin{thebibliography}{10}\itemsep=-1pt

\bibitem{uvit}
Fan Bao, Chongxuan Li, Yue Cao, and Jun Zhu.
\newblock {All are Worth Words: a ViT Backbone for Score-based Diffusion
  Models}.
\newblock {\em arXiv preprint arXiv:2209.12152}, 2022.

\bibitem{evaluations}
Ali Borji.
\newblock {Pros and Cons of GAN Evaluation Measures: New Developments}.
\newblock {\em Computer Vision and Image Understanding}, 215:103329, 2022.

\bibitem{biggan}
Andrew Brock, Jeff Donahue, and Karen Simonyan.
\newblock {Large Scale GAN Training for High Fidelity Natural Image Synthesis}.
\newblock {\em arXiv preprint arXiv:1809.11096}, 2018.

\bibitem{detr}
Nicolas Carion, Francisco Massa, Gabriel Synnaeve, Nicolas Usunier, Alexander
  Kirillov, and Sergey Zagoruyko.
\newblock {End-to-End Object Detection with Transformers}.
\newblock In {\em In Proceedings of the European Conference on Computer
  Vision}, pages 213--229. Springer, 2020.

\bibitem{cc12m}
Soravit Changpinyo, Piyush Sharma, Nan Ding, and Radu Soricut.
\newblock {Conceptual 12M: Pushing Web-Scale Image-Text Pre-Training To
  Recognize Long-Tail Visual Concepts}.
\newblock In {\em Proceedings of the IEEE/CVF Conference on Computer Vision and
  Pattern Recognition}, pages 3558--3568, 2021.

\bibitem{p2loss}
Jooyoung Choi, Jungbeom Lee, Chaehun Shin, Sungwon Kim, Hyunwoo Kim, and
  Sungroh Yoon.
\newblock {Perception Prioritized Training of Diffusion Models}.
\newblock In {\em Proceedings of the IEEE/CVF Conference on Computer Vision and
  Pattern Recognition}, pages 11472--11481, 2022.

\bibitem{fid}
Min~Jin Chong and David Forsyth.
\newblock {Effectively Unbiased FID and Inception Score and where to find
  them}.
\newblock In {\em Proceedings of the IEEE/CVF Conference on Computer Vision and
  Pattern Recognition}, pages 6070--6079, 2020.

\bibitem{coatnet}
Zihang Dai, Hanxiao Liu, Quoc~V Le, and Mingxing Tan.
\newblock {CoAtNet: Marrying Convolution and Attention for All Data Sizes}.
\newblock {\em Advances in Neural Information Processing Systems},
  34:3965--3977, 2021.

\bibitem{beatsgan}
Prafulla Dhariwal and Alexander Nichol.
\newblock {Diffusion Models Beat GANs on Image Synthesis}.
\newblock {\em {Advances in Neural Information Processing Systems}},
  34:8780--8794, 2021.

\bibitem{vit}
Alexey Dosovitskiy, Lucas Beyer, Alexander Kolesnikov, Dirk Weissenborn,
  Xiaohua Zhai, Thomas Unterthiner, Mostafa Dehghani, Matthias Minderer, Georg
  Heigold, Sylvain Gelly, Jakob Uszkoreit, and Neil Houlsby.
\newblock {An Image is Worth 16x16 Words: Transformers for Image Recognition at
  Scale}.
\newblock In {\em International Conference on Learning Representations}, 2021.

\bibitem{empirical}
Zi-Yi Dou, Yichong Xu, Zhe Gan, Jianfeng Wang, Shuohang Wang, Lijuan Wang,
  Chenguang Zhu, Pengchuan Zhang, Lu Yuan, Nanyun Peng, et~al.
\newblock {An Empirical Study of Training End-to-End Vision-and-Language
  Transformers}.
\newblock In {\em Proceedings of the IEEE/CVF Conference on Computer Vision and
  Pattern Recognition}, pages 18166--18176, 2022.

\bibitem{convit}
St{\'e}phane d’Ascoli, Hugo Touvron, Matthew~L Leavitt, Ari~S Morcos, Giulio
  Biroli, and Levent Sagun.
\newblock {ConViT: Improving Vision Transformers with Soft Convolutional
  Inductive Biases}.
\newblock In {\em International Conference on Machine Learning}, pages
  2286--2296. PMLR, 2021.

\bibitem{mvit}
Haoqi Fan, Bo Xiong, Karttikeya Mangalam, Yanghao Li, Zhicheng Yan, Jitendra
  Malik, and Christoph Feichtenhofer.
\newblock {Multiscale Vision Transformers}.
\newblock In {\em Proceedings of the IEEE/CVF International Conference on
  Computer Vision}, pages 6824--6835, 2021.

\bibitem{DBLP:journals/corr/abs-2104-12753}
Chengyue Gong, Dilin Wang, Meng Li, Vikas Chandra, and Qiang Liu.
\newblock {Improve Vision Transformers Training by Suppressing Over-smoothing}.
\newblock {\em CoRR}, abs/2104.12753, 2021.

\bibitem{he2022masked}
Kaiming He, Xinlei Chen, Saining Xie, Yanghao Li, Piotr Doll{\'a}r, and Ross
  Girshick.
\newblock {Masked Autoencoders Are Scalable Vision Learners}.
\newblock In {\em Proceedings of the IEEE/CVF Conference on Computer Vision and
  Pattern Recognition}, pages 16000--16009, 2022.

\bibitem{he2016deep}
Kaiming He, Xiangyu Zhang, Shaoqing Ren, and Jian Sun.
\newblock {Deep Residual Learning for Image Recognition}.
\newblock In {\em Proceedings of the IEEE Conference on Computer Vision and
  Pattern Recognition}, pages 770--778, 2016.

\bibitem{TTUR}
Martin Heusel, Hubert Ramsauer, Thomas Unterthiner, Bernhard Nessler,
  G{\"{u}}nter Klambauer, and Sepp Hochreiter.
\newblock {GANs Trained by a Two Time-Scale Update Rule Converge to a Nash
  Equilibrium}.
\newblock {\em CoRR}, abs/1706.08500, 2017.

\bibitem{imagen_video}
Jonathan Ho, William Chan, Chitwan Saharia, Jay Whang, Ruiqi Gao, Alexey
  Gritsenko, Diederik~P Kingma, Ben Poole, Mohammad Norouzi, David~J Fleet,
  et~al.
\newblock {Imagen Video: High Definition Video Generation with Diffusion
  Models}.
\newblock {\em arXiv preprint arXiv:2210.02303}, 2022.

\bibitem{ddpm}
Jonathan Ho, Ajay Jain, and Pieter Abbeel.
\newblock {Denoising Diffusion Probabilistic Models}.
\newblock {\em {Advances in Neural Information Processing Systems}},
  33:6840--6851, 2020.

\bibitem{classifier-free}
Jonathan Ho and Tim Salimans.
\newblock {Classifier-free Diffusion Guidance}.
\newblock {\em arXiv preprint arXiv:2207.12598}, 2022.

\bibitem{transgan}
Yifan Jiang, Shiyu Chang, and Zhangyang Wang.
\newblock {TransGAN: Two Pure Transformers Can Make One Strong GAN, and That
  Can Scale Up}.
\newblock {\em arXiv preprint arXiv:2102.07074}, 1(3), 2021.

\bibitem{stylegan2}
Tero Karras, Miika Aittala, Janne Hellsten, Samuli Laine, Jaakko Lehtinen, and
  Timo Aila.
\newblock {Training Generative Adversarial Networks with Limited Data}.
\newblock {\em Advances in Neural Information Processing Systems},
  33:12104--12114, 2020.

\bibitem{kim2022soft}
Dongjun Kim, Seungjae Shin, Kyungwoo Song, Wanmo Kang, and Il-Chul Moon.
\newblock {Soft Truncation: A Universal Training Technique of Score-based
  Diffusion Model for High Precision Score Estimation}.
\newblock In {\em International Conference on Machine Learning}, pages
  11201--11228. PMLR, 2022.

\bibitem{vilt}
Wonjae Kim, Bokyung Son, and Ildoo Kim.
\newblock {ViLT: Vision-and-Language Transformer Without Convolution or Region
  Supervision}.
\newblock In {\em International Conference on Machine Learning}, pages
  5583--5594. PMLR, 2021.

\bibitem{vae}
Diederik~P. Kingma and Max Welling.
\newblock {Auto-Encoding Variational Bayes}.
\newblock In {\em International Conference on Learning Representations}, 2014.

\bibitem{vitgan}
Kwonjoon Lee, Huiwen Chang, Lu Jiang, Han Zhang, Zhuowen Tu, and Ce Liu.
\newblock {ViTGAN: Training GANs with Vision Transformers}.
\newblock {\em arXiv preprint arXiv:2107.04589}, 2021.

\bibitem{dn_detr}
Feng Li, Hao Zhang, Shilong Liu, Jian Guo, Lionel~M Ni, and Lei Zhang.
\newblock {DN-DETR: Accelerate DETR Training by Introducing Query DeNoising}.
\newblock In {\em Proceedings of the IEEE/CVF Conference on Computer Vision and
  Pattern Recognition}, pages 13619--13627, 2022.

\bibitem{swinv2imagen}
Ruijun Li, Weihua Li, Yi Yang, Hanyu Wei, Jianhua Jiang, and Quan Bai.
\newblock {Swinv2-Imagen: Hierarchical Vision Transformer Diffusion Models for
  Text-to-Image Generation}.
\newblock {\em arXiv preprint arXiv:2210.09549}, 2022.

\bibitem{dab}
Shilong Liu, Feng Li, Hao Zhang, Xiao Yang, Xianbiao Qi, Hang Su, Jun Zhu, and
  Lei Zhang.
\newblock {DAB-DETR: Dynamic Anchor Boxes are Better Queries for DETR}.
\newblock In {\em International Conference on Learning Representations}, 2022.

\bibitem{swin}
Ze Liu, Yutong Lin, Yue Cao, Han Hu, Yixuan Wei, Zheng Zhang, Stephen Lin, and
  Baining Guo.
\newblock {Swin Transformer: Hierarchical Vision Transformer using Shifted
  Windows}.
\newblock In {\em Proceedings of the IEEE/CVF International Conference on
  Computer Vision}, pages 10012--10022, 2021.

\bibitem{celeba}
Ziwei Liu, Ping Luo, Xiaogang Wang, and Xiaoou Tang.
\newblock {Deep Learning Face Attributes in the Wild}.
\newblock In {\em Proceedings of the IEEE international conference on computer
  vision}, pages 3730--3738, 2015.

\bibitem{FCN}
Jonathan Long, Evan Shelhamer, and Trevor Darrell.
\newblock {Fully Convolutional Networks for Semantic Segmentation}.
\newblock In {\em Proceedings of the IEEE Conference on Computer Vision and
  Pattern Recognition}, pages 3431--3440, 2015.

\bibitem{adamw}
Ilya Loshchilov and Frank Hutter.
\newblock {Decoupled Weight Decay Regularization}.
\newblock {\em arXiv preprint arXiv:1711.05101}, 2017.

\bibitem{meng2022distillation}
Chenlin Meng, Ruiqi Gao, Diederik~P Kingma, Stefano Ermon, Jonathan Ho, and Tim
  Salimans.
\newblock {On Distillation of Guided Diffusion Models}.
\newblock {\em arXiv preprint arXiv:2210.03142}, 2022.

\bibitem{sdedit}
Chenlin Meng, Yutong He, Yang Song, Jiaming Song, Jiajun Wu, Jun-Yan Zhu, and
  Stefano Ermon.
\newblock {SDEdit: Guided Image Synthesis and Editing with Stochastic
  Differential Equations}.
\newblock In {\em International Conference on Learning Representations}, 2021.

\bibitem{glide}
Alex Nichol, Prafulla Dhariwal, Aditya Ramesh, Pranav Shyam, Pamela Mishkin,
  Bob McGrew, Ilya Sutskever, and Mark Chen.
\newblock {GLIDE: Towards Photorealistic Image Generation and Editing with
  Text-Guided Diffusion Models}.
\newblock {\em arXiv preprint arXiv:2112.10741}, 2021.

\bibitem{improved_ddpm}
Alexander~Quinn Nichol and Prafulla Dhariwal.
\newblock {Improved Denoising Diffusion Probabilistic Models}.
\newblock In {\em International Conference on Machine Learning}, pages
  8162--8171. PMLR, 2021.

\bibitem{DiTs}
William Peebles and Saining Xie.
\newblock {Scalable Diffusion Models with Transformers}.
\newblock {\em arXiv preprint arXiv:2212.09748}, 2022.

\bibitem{CLIP}
Alec Radford, Jong~Wook Kim, Chris Hallacy, Aditya Ramesh, Gabriel Goh,
  Sandhini Agarwal, Girish Sastry, Amanda Askell, Pamela Mishkin, Jack Clark,
  et~al.
\newblock {Learning Transferable Visual Models From Natural Language
  Supervision}.
\newblock In {\em International Conference on Machine Learning}, pages
  8748--8763. PMLR, 2021.

\bibitem{raffel2020t5}
Colin Raffel, Noam Shazeer, Adam Roberts, Katherine Lee, Sharan Narang, Michael
  Matena, Yanqi Zhou, Wei Li, Peter~J Liu, et~al.
\newblock {Exploring the Limits of Transfer Learning with a Unified
  Text-to-Text Transformer}.
\newblock {\em J. Mach. Learn. Res.}, 21(140):1--67, 2020.

\bibitem{dalle2}
Aditya Ramesh, Prafulla Dhariwal, Alex Nichol, Casey Chu, and Mark Chen.
\newblock {Hierarchical Text-Conditional Image Generation with CLIP Latents}.
\newblock {\em arXiv preprint arXiv:2204.06125}, 2022.

\bibitem{latentdiffusion}
Robin Rombach, Andreas Blattmann, Dominik Lorenz, Patrick Esser, and Bj{\"o}rn
  Ommer.
\newblock {High-Resolution Image Synthesis with Latent Diffusion Models}.
\newblock In {\em Proceedings of the IEEE/CVF Conference on Computer Vision and
  Pattern Recognition}, pages 10684--10695, 2022.

\bibitem{palette}
Chitwan Saharia, William Chan, Huiwen Chang, Chris Lee, Jonathan Ho, Tim
  Salimans, David Fleet, and Mohammad Norouzi.
\newblock {Palette: Image-to-Image Diffusion Models}.
\newblock In {\em ACM SIGGRAPH 2022 Conference Proceedings}, pages 1--10, 2022.

\bibitem{imagen}
Chitwan Saharia, William Chan, Saurabh Saxena, Lala Li, Jay Whang, Emily
  Denton, Seyed Kamyar~Seyed Ghasemipour, Burcu~Karagol Ayan, S~Sara Mahdavi,
  Rapha~Gontijo Lopes, et~al.
\newblock {Photorealistic Text-to-Image Diffusion Models with Deep Language
  Understanding}.
\newblock {\em arXiv preprint arXiv:2205.11487}, 2022.

\bibitem{sr3}
Chitwan Saharia, Jonathan Ho, William Chan, Tim Salimans, David~J Fleet, and
  Mohammad Norouzi.
\newblock {Image Super-Resolution via Iterative Refinement}.
\newblock {\em IEEE Transactions on Pattern Analysis and Machine Intelligence},
  2022.

\bibitem{progressive}
Tim Salimans and Jonathan Ho.
\newblock {Progressive Distillation for Fast Sampling of Diffusion Models}.
\newblock {\em arXiv preprint arXiv:2202.00512}, 2022.

\bibitem{pixelcnn_plus_plus}
Tim Salimans, Andrej Karpathy, Xi Chen, and Diederik~P. Kingma.
\newblock {PixelCNN++: Improving the PixelCNN with Discretized Logistic Mixture
  Likelihood and Other Modifications}.
\newblock In {\em International Conference on Learning Representations}, 2017.

\bibitem{laion5b}
Christoph Schuhmann, Romain Beaumont, Richard Vencu, Cade Gordon, Ross
  Wightman, Mehdi Cherti, Theo Coombes, Aarush Katta, Clayton Mullis, Mitchell
  Wortsman, et~al.
\newblock {LAION-5B: An open large-scale dataset for training next generation
  image-text models}.
\newblock {\em arXiv preprint arXiv:2210.08402}, 2022.

\bibitem{sohl2015deep}
Jascha Sohl-Dickstein, Eric Weiss, Niru Maheswaranathan, and Surya Ganguli.
\newblock {Deep Unsupervised Learning using Nonequilibrium Thermodynamics}.
\newblock In {\em International Conference on Machine Learning}, pages
  2256--2265. PMLR, 2015.

\bibitem{song2021denoising}
Jiaming Song, Chenlin Meng, and Stefano Ermon.
\newblock {Denoising Diffusion Implicit Models}.
\newblock In {\em International Conference on Learning Representations}, 2021.

\bibitem{ncsn}
Yang Song and Stefano Ermon.
\newblock {Generative Modeling by Estimating Gradients of the Data
  Distribution}.
\newblock {\em Advances in Neural Information Processing Systems}, 32, 2019.

\bibitem{ncsnv2}
Yang Song and Stefano Ermon.
\newblock {Improved Techniques for Training Score-Based Generative Models}.
\newblock {\em Advances in Neural Information Processing Systems},
  33:12438--12448, 2020.

\bibitem{song2021scorebased}
Yang Song, Jascha Sohl-Dickstein, Diederik~P Kingma, Abhishek Kumar, Stefano
  Ermon, and Ben Poole.
\newblock {Score-Based Generative Modeling through Stochastic Differential
  Equations}.
\newblock In {\em International Conference on Learning Representations}, 2021.

\bibitem{segformer}
Robin Strudel, Ricardo Garcia, Ivan Laptev, and Cordelia Schmid.
\newblock {Segmenter: Transformer for Semantic Segmentation}.
\newblock In {\em Proceedings of the IEEE/CVF International Conference on
  Computer Vision}, pages 7262--7272, 2021.

\bibitem{deit}
Hugo Touvron, Matthieu Cord, Matthijs Douze, Francisco Massa, Alexandre
  Sablayrolles, and Herv{\'e} J{\'e}gou.
\newblock {Training data-efficient image transformers \& distillation through
  attention}.
\newblock In {\em International Conference on Machine Learning}, pages
  10347--10357. PMLR, 2021.

\bibitem{cub-bird}
Catherine Wah, Steve Branson, Peter Welinder, Pietro Perona, and Serge
  Belongie.
\newblock {The Caltech-UCSD Birds-200-2011 Dataset}.
\newblock 2011.

\bibitem{pvt}
Wenhai Wang, Enze Xie, Xiang Li, Deng-Ping Fan, Kaitao Song, Ding Liang, Tong
  Lu, Ping Luo, and Ling Shao.
\newblock {Pyramid Vision Transformer: A Versatile Backbone for Dense
  Prediction without Convolutions}.
\newblock In {\em Proceedings of the IEEE/CVF International Conference on
  Computer Vision}, pages 568--578, 2021.

\bibitem{cvt}
Haiping Wu, Bin Xiao, Noel Codella, Mengchen Liu, Xiyang Dai, Lu Yuan, and Lei
  Zhang.
\newblock {CvT: Introducing Convolutions to Vision Transformers}.
\newblock In {\em Proceedings of the IEEE/CVF International Conference on
  Computer Vision}, pages 22--31, 2021.

\bibitem{group_norm}
Yuxin Wu and Kaiming He.
\newblock {Group Normalization}.
\newblock In {\em Proceedings of the European Conference on Computer Vision
  (ECCV)}, pages 3--19, 2018.

\bibitem{ddgan}
Zhisheng Xiao, Karsten Kreis, and Arash Vahdat.
\newblock Tackling the generative learning trilemma with denoising diffusion
  {GAN}s.
\newblock In {\em International Conference on Learning Representations}, 2022.

\bibitem{probing}
Hongwei Xue, Yupan Huang, Bei Liu, Houwen Peng, Jianlong Fu, Houqiang Li, and
  Jiebo Luo.
\newblock {Probing Inter-modality: Visual Parsing with Self-Attention for
  Vision-Language Pre-training}.
\newblock {\em Advances in Neural Information Processing Systems},
  34:4514--4528, 2021.

\bibitem{genvit}
Xiulong Yang, Sheng-Min Shih, Yinlin Fu, Xiaoting Zhao, and Shihao Ji.
\newblock {Your ViT is Secretly a Hybrid Discriminative-Generative Diffusion
  Model}.
\newblock {\em arXiv preprint arXiv:2208.07791}, 2022.

\bibitem{lsun}
Fisher Yu, Ari Seff, Yinda Zhang, Shuran Song, Thomas Funkhouser, and Jianxiong
  Xiao.
\newblock {LSUN: Construction of a Large-scale Image Dataset using Deep
  Learning with Humans in the Loop}.
\newblock {\em arXiv preprint arXiv:1506.03365}, 2015.

\bibitem{dino}
Hao Zhang, Feng Li, Shilong Liu, Lei Zhang, Hang Su, Jun Zhu, Lionel~M Ni, and
  Heung-Yeung Shum.
\newblock {DINO: DETR with Improved DeNoising Anchor Boxes for End-to-End
  Object Detection}.
\newblock {\em arXiv preprint arXiv:2203.03605}, 2022.

\bibitem{NesT}
Zizhao Zhang, Han Zhang, Long Zhao, Ting Chen, Sercan~{\"O} Arik, and Tomas
  Pfister.
\newblock {Nested Hierarchical Transformer: Towards Accurate, Data-Efficient
  and Interpretable Visual Understanding}.
\newblock In {\em Proceedings of the AAAI Conference on Artificial
  Intelligence}, volume~36, pages 3417--3425, 2022.

\bibitem{rethinking_segmentation}
Sixiao Zheng, Jiachen Lu, Hengshuang Zhao, Xiatian Zhu, Zekun Luo, Yabiao Wang,
  Yanwei Fu, Jianfeng Feng, Tao Xiang, Philip~HS Torr, et~al.
\newblock {Rethinking Semantic Segmentation from a Sequence-to-Sequence
  Perspective with Transformers}.
\newblock In {\em Proceedings of the IEEE/CVF Conference on Computer Vision and
  Pattern Recognition}, pages 6881--6890, 2021.

\bibitem{trar}
Yiyi Zhou, Tianhe Ren, Chaoyang Zhu, Xiaoshuai Sun, Jianzhuang Liu, Xinghao
  Ding, Mingliang Xu, and Rongrong Ji.
\newblock {TRAR: Routing the Attention Spans in Transformer for Visual Question
  Answering}.
\newblock In {\em Proceedings of the IEEE/CVF International Conference on
  Computer Vision}, pages 2074--2084, 2021.

\end{thebibliography}
}

\newpage
\appendix
\label{sec:appendix}
\section*{Supplementary Material}

\section{Additional Details on Conditional Diffusion Models}
A conditional diffusion model conditions the backward process with external information. Formally, $\textbf{c}$ denotes conditional information (e.g. category label or text prompt), and the new joint distribution conditional on $\textbf{c}$ can be written as:
\begin{eqnarray}
& p_{\theta}\left(\mathbf{x}_{0: T} \mid \mathbf{c}\right)=p\left(\mathbf{x}_{T}\right) \prod_{t=1}^{T} p_{\theta}\left(\mathbf{x}_{t-1} \mid \mathbf{x}_{t}, \mathbf{c}\right)
\end{eqnarray}
where
\begin{eqnarray}
& p_{\theta}\left(\mathbf{x}_{t-1} \mid \mathbf{x}_{t}, \mathbf{c}\right)=\mathcal{N}\left(\mathbf{x}_{t-1} ; \mu_{\theta}\left(\mathbf{x}_{t}, t, \mathbf{c}\right), \Sigma_{\theta}\left(\mathbf{x}_{t}, t, \mathbf{c}\right)\right) \nonumber
\end{eqnarray}

In practice, a conditional diffusion model is usually supplemented with gradient information either from a pre-trained discriminative model (classifier for label-conditional\cite{beatsgan} and CLIP for text-conditional\cite{glide}) or classifier-free guidance\cite{classifier-free} via jointly training diffusion model with and without external conditioning.

\paragraph{Classifier Guidance} Dhariwal \etal\cite{beatsgan} find that with classifier guidance, samples from class-conditional diffusion models may often be improved. The main idea of classifier guidance is to use a trained classifier $p(\mathbf{c}|\mathbf{x}_t)$ as supervisor to provide gradient guidance, mixed with the original score during sampling. Specifically, during sampling we use a modified score $\nabla_{\mathbf{x}_{t}}\left[\log p\left(\mathbf{x}_{t} \mid \mathbf{c}\right)+\omega \log p\left(\mathbf{c} \mid \mathbf{x}_{t}\right)\right]$ to approximate samples from the distribution $\tilde{p}\left(\mathbf{x}_{t} \mid \mathbf{c}\right) \propto p\left(\mathbf{x}_{t} \mid \mathbf{c}\right) p\left(\mathbf{c} \mid \mathbf{x}_{t}\right)^{\omega}$, where $\omega$ denotes the guidance scale.

\paragraph{CLIP Guidance} Radford \etal\cite{CLIP} propose CLIP as a scalable method for learning representations of texts and images, encouraging paired texts and images to have higher similarity in latent space. Since CLIP can evaluate how close an image is to a caption, GLIDE\cite{glide} proposes to implement text-to-image synthesis by introducing CLIP as a tool to steer generation. GLIDE replaces the classifier with a CLIP model as in classifier guidance, using the gradient of the dot product of the caption and image encodings with regard to the image to perturb the reverse process score model: $\nabla_{\mathbf{x}_{t}}\left[\log p\left(\mathbf{x}_{t} \mid \mathbf{c}\right)+\omega \left(f(\mathbf{x})\cdot g(\mathbf{c})\right)\right]$. Additionally, CLIP needs to be trained on noisy images $\mathbf{x}_t$ to obtain the correct score estimation on the noised inputs. Through experiments, the CLIP-guided model shows better generative performance.

\paragraph{Classifier-free Guidance} Unfortunately, few snags make the classifier-guided diffusion impractical. First, because the diffusion models operate by gradually denoising inputs, any classifier used for guidance also needs to be able to cope with different levels of noised data, requiring training of a bespoke classifier specifically for guidance. Next, even with a noise-robust classifier, classifier guidance is inherently limited in its effectiveness: most of the information in the input $\mathbf{x}_t$ is irrelevant to predicting label $y$, so the gradient adjustment made by the classifier alone is hardly an informative guidance in input space. Ho \etal\cite{classifier-free} propose classifier-free guidance, a technique that guides diffusion models without requiring a separate classifier model to be trained. It designs an implicit classifier by jointly training a conditional and unconditional diffusion model. Specifically, one trains a conditional diffusion model $\epsilon_\theta(\mathbf{x}_t|y)$, with conditioning dropout: with predefined probability, the conditioning information $y$ is dropped (in practice, $y$ is often replaced with a special $blank$ input value $\emptyset$ denoting the absence of conditioning information). During sampling, the output of the model is extrapolated further in the direction of $\epsilon_\theta(\mathbf{x}_t|y)$ And away from $\epsilon_\theta(\mathbf{x}_t|\emptyset)$ as follows:
\begin{equation}
\begin{aligned}
\hat{\epsilon}_{\theta}\left(\mathbf{x}_{t} \mid y\right)=\epsilon_{\theta}\left(\mathbf{x} \mid \emptyset\right)+s \cdot\left(\epsilon_{\theta}\left(\mathbf{x} \mid y\right)-\epsilon_{\theta}\left(\mathbf{x} \mid \emptyset\right)\right)
\end{aligned}
\end{equation}
Here $s\geq1$ is denoted as the guidance scale. The classifier-free method enables a single model to rely on its own knowledge for guidance instead of depending on a separate discriminative model. 

\section{Implementation Details}
\label{appendix:implement}
\subsection{IU-ViT Model details}
\begin{figure*}[h]
\centering
\includegraphics[width=0.95\linewidth]{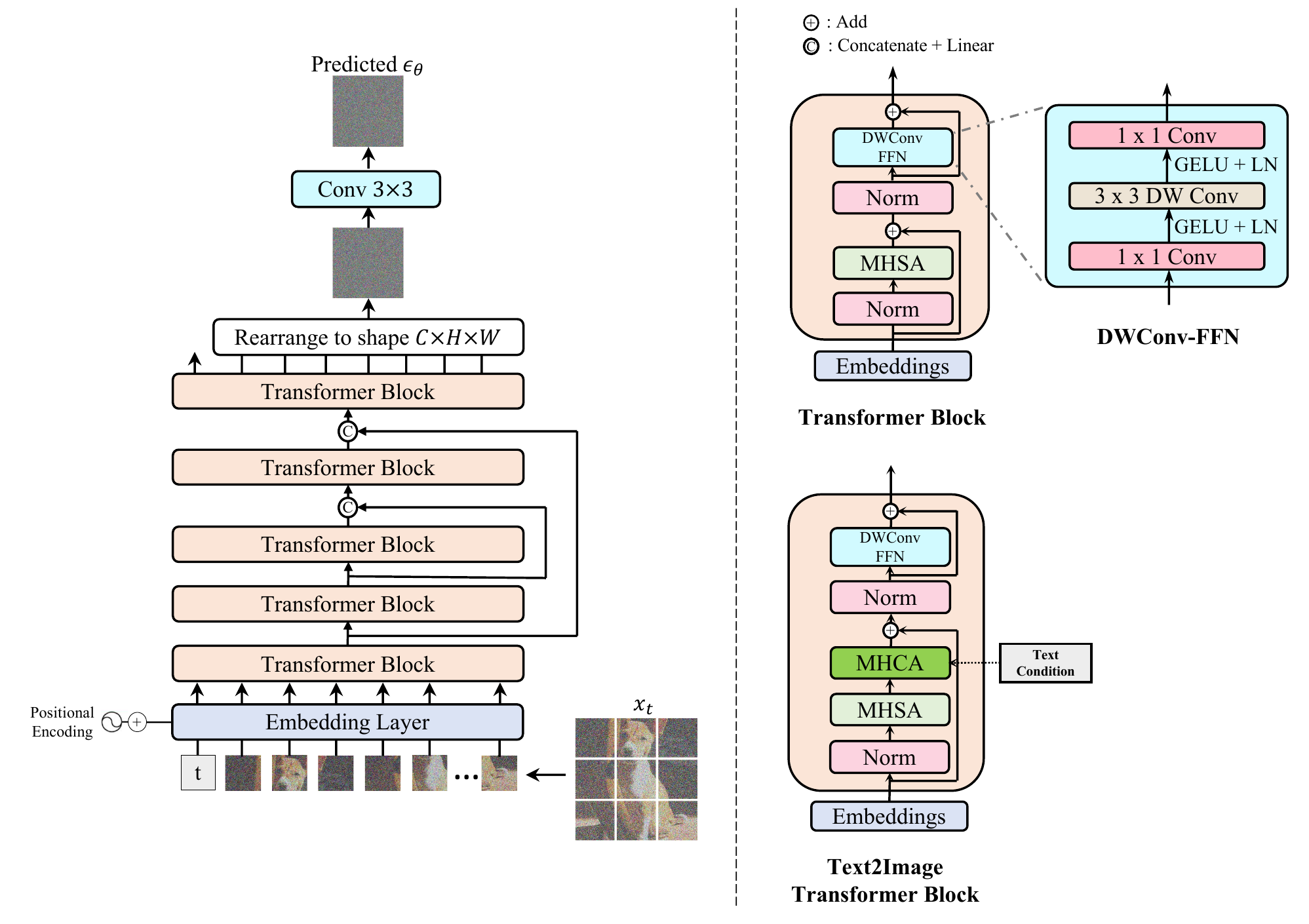}
\caption{The IU-ViT architecture. \textit{MHSA}: Multi-head self-attention, \textit{MHCA}: Multi-head cross-attention.}
\label{fig:iuvit_arch}
\end{figure*}

 Our IU-ViT is based on previous work of U-ViT\cite{uvit} which uses \textit{patch embedding} to project pixels into image tokens then prepend a time-embedding token as inputs to a stack of transformers. The time token is stripped after the final transformer layer and the image tokens are projected and reshaped to produce the final noise predictions. Similar to U-ViT, we concatenate upper transformer layer features with lower layer features in the channel dimension via skip connections and then project back to the original channel size for computation efficiency. The overall architecture is illustrated in Figure \ref{fig:iuvit_arch}.

We first evaluate IU-ViT on CIFAR-10. For this 32$\times$32 resolution task, we use a 13-layer ViT of size 45M (on par with U-ViT) with two simple improvements as detailed in \ref{iuvit}. All experiments are conducted on 4 NVIDIA A100(40G) GPUs with per GPU batch size of 128. In the computation of the FID, we sampled 50,000 images as the TTUR \cite{TTUR} repository suggested and achieve the best FID result (2.56) on the ViT-based backbone, as shown in Table \ref{cifar10-generation-results}. We also evaluate IU-ViT on CelebA $64\times64$ with similar network hyperparameters as U-ViT suggested and achieve a \textcolor{black}{new} \textbf{state-of-the-art FID result (1.57)}.

\begin{figure*}[h]
\centering
\includegraphics[width=0.95\linewidth]{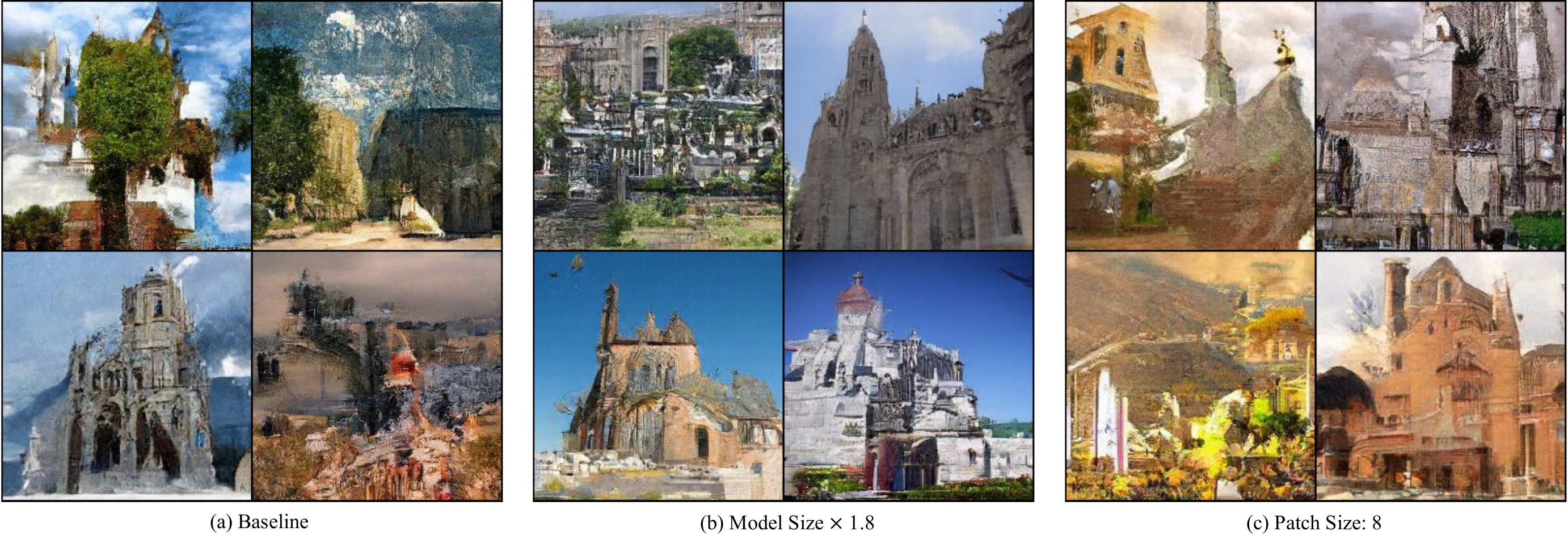}
\caption{IU-ViT $256\times256$ LSUN Church model variations. (a) Baseline model, details provided in Table \ref{tab:iuvit-hyper}; (b) Expand Transformer block \textit{hidden\_size} from 1536 to 2048, model parameter size from 527M to 935M; (c) Reduce \textit{patch\_size} from 16 to 8. Neither improvements is able to sufficiently boost generation quality at the expense of practicality.}
\label{fig:iuvit_highres_ablation}
\end{figure*}

We provide detailed experimental settings in Table \ref{tab:iuvit-hyper} for higher-resolution image synthesis tasks.
In addition, we explored whether we could break the bottleneck of high-resolution image synthesis by increasing the transformer block hidden size and reducing the patch size, but found that scaling up model size or sacrificing computation efficiency to reduce the patch size \emph{could not} improve model performance. See Figure \ref{fig:iuvit_highres_ablation} for illustration.

\begin{figure*}[h]
\centering
\includegraphics[width=0.95\linewidth]{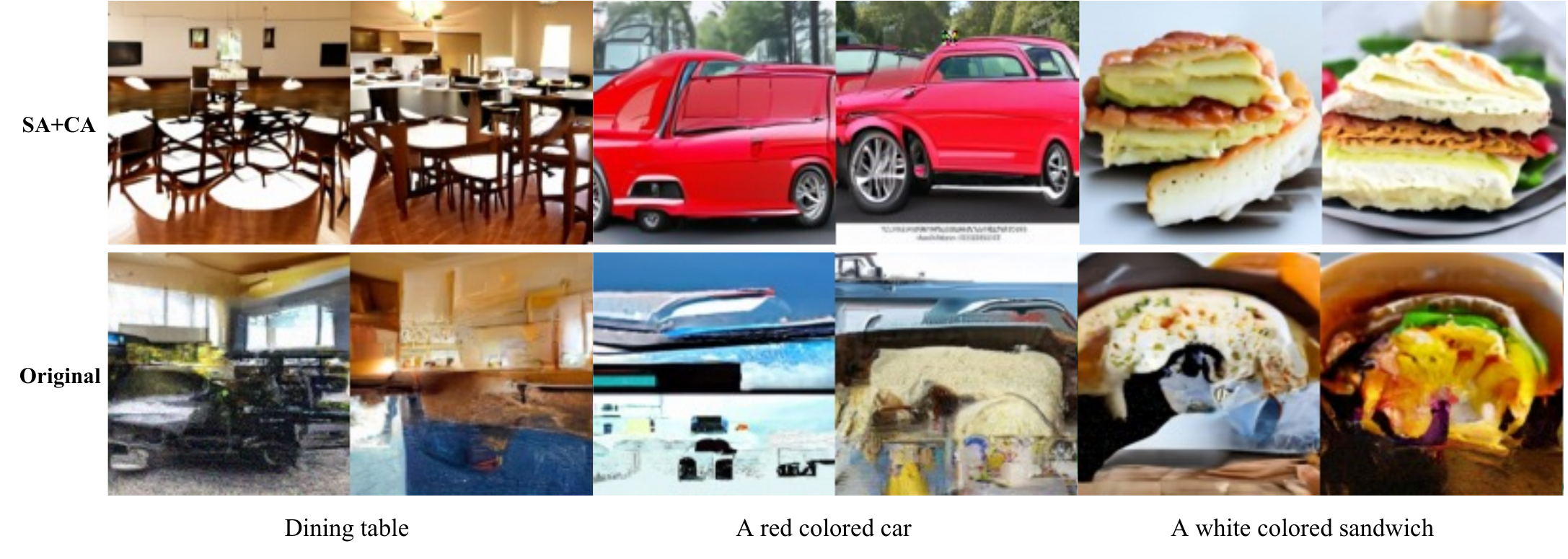}
\caption{IU-ViT $128\times128$ text-to-image comparison of different attention mechanisms. Top: separated self-attention and cross-attention; Bottom: fused "self"-attention\cite{imagen,glide,dalle2}. It can be observed that separated attention encourages better semantic alignment and instance generation.}
\label{fig:iuvit_attention_ablation}
\end{figure*}

For the text-to-image task, we use pretrained T5-XXL model for text encoding similarly to Imagen \cite{imagen}. Following Imagen's approach, the network conditions on text via a pooled text encoding vector which is concatenated with timestep embedding, it also conditions on the entire sequence of text encoding via cross-attention. In our implementation, we dedicate separate self-attention and cross-attention computations into a transformer block and also introduce the DWConv-FFN module for better localization. The effectiveness of separated attention is illustrated in Figure \ref{fig:iuvit_attention_ablation}. We compare a $64\times64$ 300M parameter text-conditional IU-ViT with a similar sized UNet-based implementation\cite{dalle2,imagen}. For more details, see Table \ref{tab:iuvit-hyper}.

\begin{table*}
\small 
    \centering
    \renewcommand\arraystretch{1.2} 
    \begin{tabular}{lccccc}
    \toprule[1.2pt]
        Dataset & CIFAR-10 & CelebA $64\times 64$ & CelebA $128\times 128$ & Church $256\times 256$ & CC12M $64\times 64$ \\ \hline
        Patch size & 2 & 4 & 8 & 16 & 4\\ 
        Layers & 13 & 13 & 17 & 17 & 17\\ 
        Hidden size & 512 & 512 & 1408 & 1536 & 1024\\ 
        Heads & 8 & 8 & 22 & 24 & 16\\ 
        Text encoder context  & - & - & - & - & 256\\
        Text encoder width  & - & - & - & - & 1024\\
        Params & 45M & 45M & 442M & 527M & 307M \\ \hline
        Diffusion steps & 1000 & 1000 & 1000 & 1000 & 1000\\ 
        Noise schedule & linear & linear & linear & linear & linear \\
        Batch size & 128 & 128 & 256 & 256 & 1024\\
        Training iterations & 500K & 500K & 450K & $150K^\dagger$ & $150k^\dagger$\\
        Optimizer & AdamW\cite{adamw} & AdamW & AdamW & AdamW & AdamW\\
        Learning rate & 2e-4 & 2e-4 & 2e-4 & 1e-4 & 1e-4\\
        EMA decay & 0.9999 & 0.9999 & 0.9999 & 0.9999 & 0.9999 \\
        Betas & (0.99, 0.999) & (0.99, 0.99) & (0.99, 0.99) & (0.99, 0.99) & (0.99, 0.99)\\ \hline
        Sampler & EM & EM & EM & EM & DDIM\cite{song2021denoising}\\
        Sampling steps & 1K & 1K & 1K & 1K &250\\
    \bottomrule[1.2pt]
    \end{tabular}
  \caption{IU-ViT experimental settings. EM represents the Euler-Maruyama sampler. $\dagger$: early stopping.}
  \label{tab:iuvit-hyper}
\end{table*}

\subsection{ASCEND Model details}
For ASCEND, we incorporate SwinTransformer Block into the encoder.  We use residual downsampling/upsampling operation to replace patch merging and patch expanding. Each SwinBlock contains 2 window-attention layers (one with window-shift and another without). For the decoder, we refer interested readers to the \textit{outblock} architecture used in \cite{beatsgan}. We provide detailed experimental settings in Table \ref{tab:ASCEND-hyper}.

\begin{table*}
\small 
    \centering
    \renewcommand\arraystretch{1.2} 
    \begin{tabular}{lcccc}
    \toprule[1.2pt]
        Resolution & $32\times 32$ & $64\times 64$ &$256\times 256$ & T2I-$128\times 128$ \\ \hline
        Channels & 128 & 192 & 256 & 192 \\
        Depth    & 3   & 2   & 2   & 3   \\
        Channels multiple & 1,2,2,2 & 1,2,3,4 & 1,1,2,2,3,4 & 1,2,3,4,4 \\ 
        Heads channels & 64 & 64 & 64 & 64 \\ 
        Text encoder context  & - & - & -  & 256\\
        Text encoder width  & - & - & -  & 1024\\
        Attention resolution & 16,8 & 32,16,8 & 32,16,8 &  64,32,16,8 \\ 
        Dropout & 0.1 & 0.1 & 0.0 & 0.0 \\\hline
        Diffusion steps & 1000 & 1000 & 1000  & 1000\\ 
        Noise schedule & cosine & cosine & cosine  & cosine \\
        Batch size & 128 & 128 & 256 & 512 \\
        Optimizer & AdamW\cite{adamw} & AdamW & AdamW & AdamW\\
        Learning rate & 1e-4 & 2e-4 & 1e-4 & 1.2e-4\\
        EMA decay & 0.9999 & 0.9999 & 0.9999  & 0.9999 \\
        Betas & (0.99, 0.999) & (0.99, 0.99)  & (0.99, 0.99) & (0.9, 0.9999)\\ \hline
        Sampler & EM & EM &  EM & DDIM\cite{song2021denoising}\\
        Sampling steps & 1K  & 1K & 1K &50\\
    \bottomrule[1.2pt]
    \end{tabular}
  \caption{ASCEND experimental settings. EM represents the Euler-Maruyama sampler. \textit{T2I}: text-to-image task}
  \label{tab:ASCEND-hyper}
\end{table*}

\section{More Visualization Results}
\label{appendix:visualization}
\subsection{IU-ViT Generation Results}

\begin{figure}[]
\centering
\includegraphics[width=0.9\linewidth]{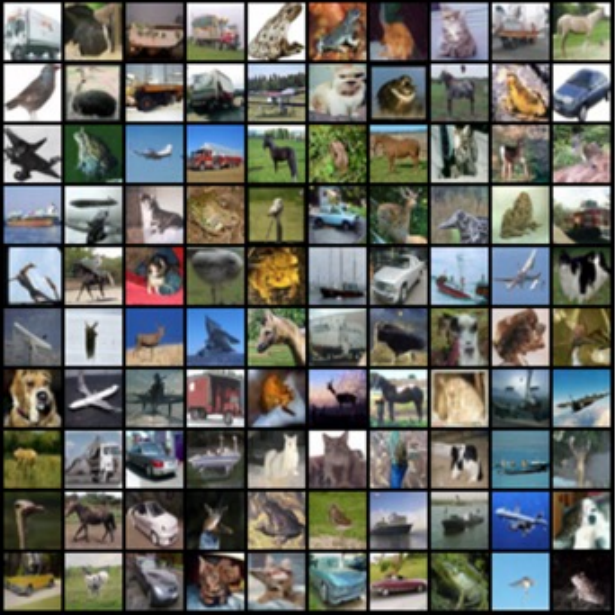}
\caption{IU-ViT: randomly sampled results on CIFAR-10 (FID=2.56), 1000 sampling steps.}
\label{fig:iuvit_cifar10}
\end{figure}

\begin{figure}[]
\centering
\includegraphics[width=0.9\linewidth]{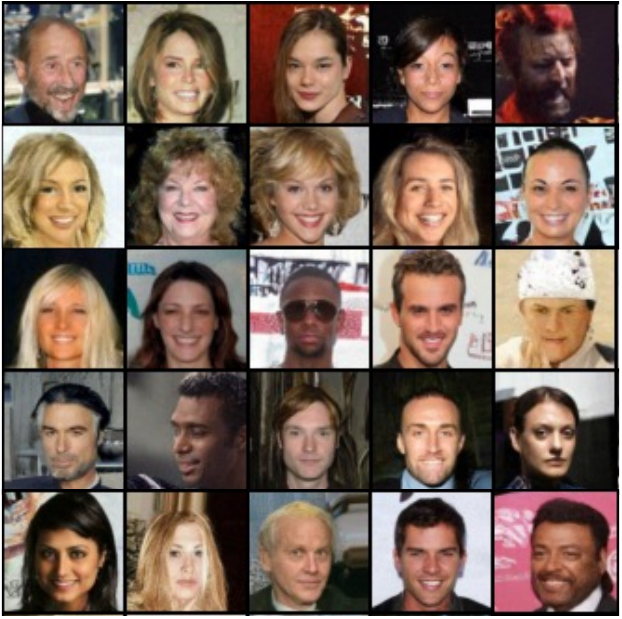}
\caption{IU-ViT: randomly sampled results on CelebA$64\times64$ (FID=1.57), 1000 sampling steps.}
\label{fig:iuvit_celeba64}
\end{figure}

\begin{figure}[]
\centering
\includegraphics[width=0.9\linewidth]{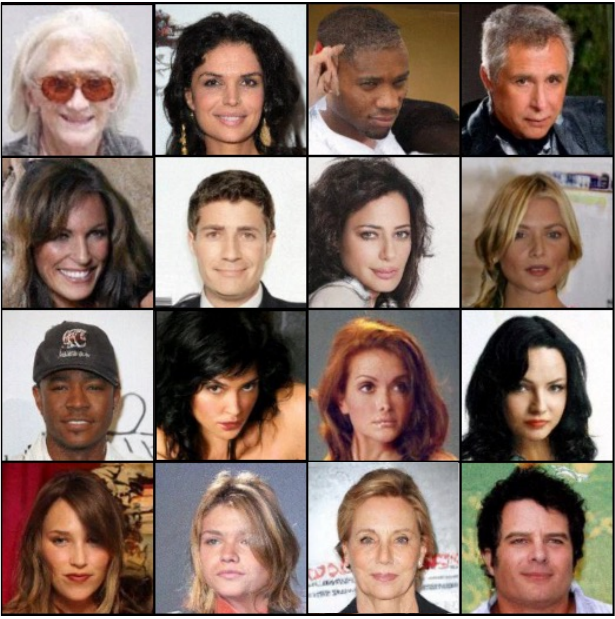}
\caption{IU-ViT: randomly sampled results on CelebA $128\times128$, 1000 sampling steps.}
\label{fig:iuvit_celeba128}
\end{figure}

\begin{figure}[]
\centering
\includegraphics[width=0.9\linewidth]{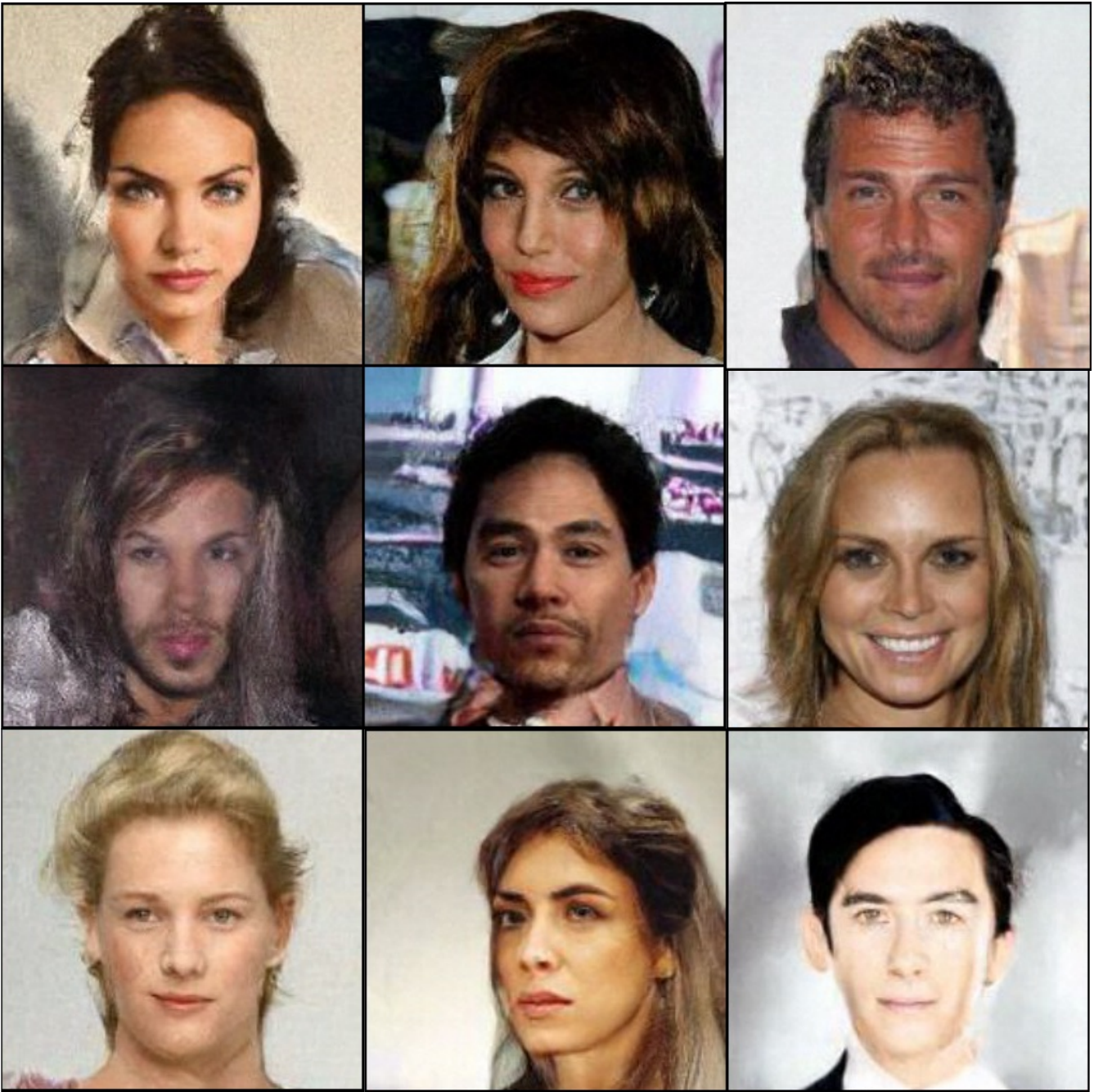}
\caption{IU-ViT: randomly sampled results on CelebA $256\times256$. The local features of the generated results are visibly blurred with evident aliases, 1000 sampling steps. }
\label{fig:iuvit_celeba256}
\end{figure}

\begin{figure*}[h]
    \centering
    \includegraphics[width=0.95\linewidth]{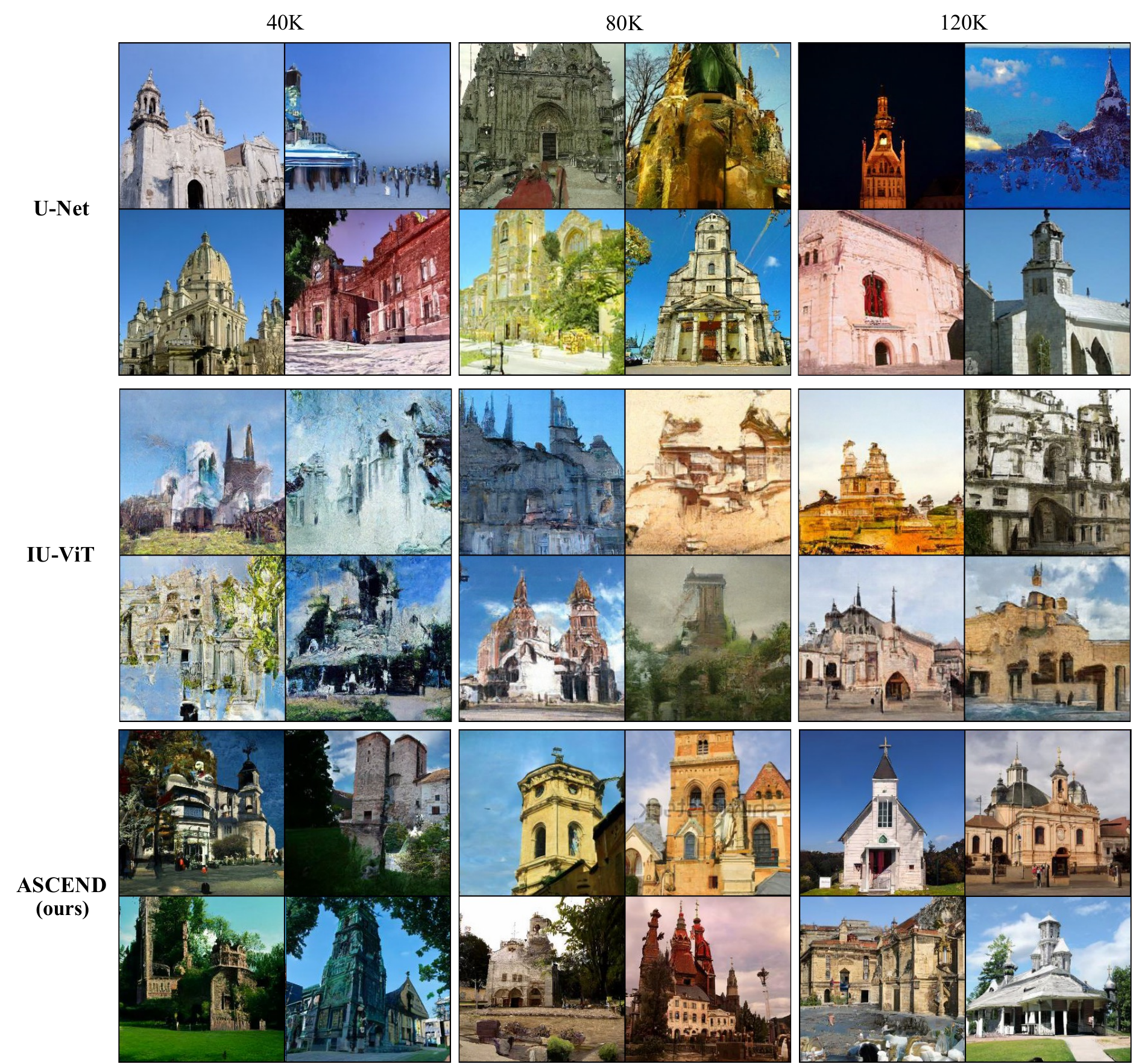}
    
    \caption{More LSUN-Church $256\times256$ samples with U-Net (top), IU-ViT (middle) and ASCEND (bottom).}
\label{fig:church256_comparison}
\end{figure*}

\begin{figure*}[h]
    \centering
    \includegraphics[width=0.95\linewidth]{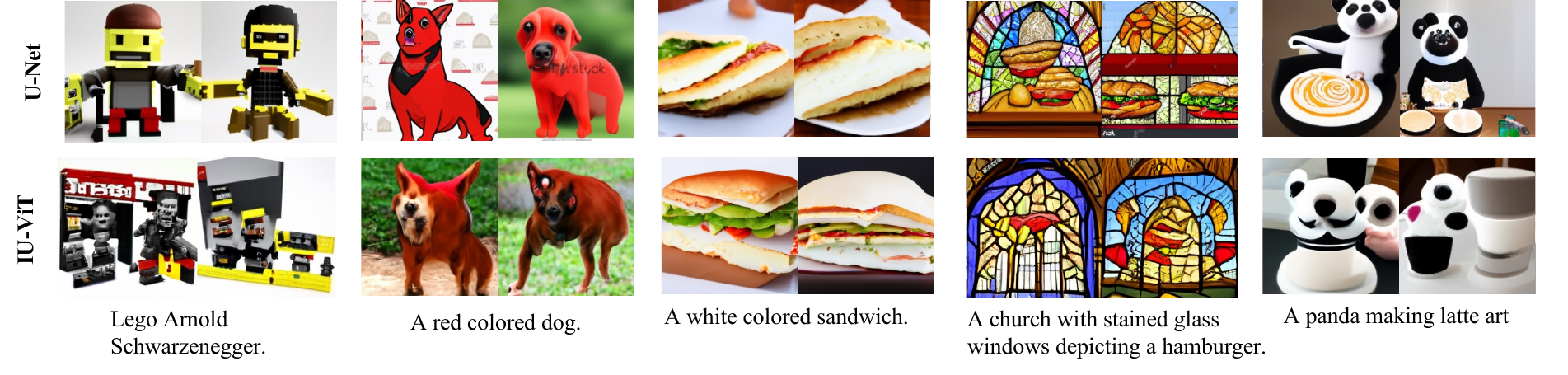}
    
    \caption{Example qualitative comparisons of text-to-image $64\times64$ models based on U-Net\cite{imagen,dalle2}  and IU-ViT, evaluated on DrawBench prompts. Both models are of similar size (300M). We observed that the IU-ViT model struggles more with synthesizing realistic shapes and natural images compared to the UNet-based model.}
\label{fig:iuvit_t2i_64}
\end{figure*}

\clearpage
\clearpage
\subsection{ASCEND Generation Results}
\begin{figure}[h]
\centering
\includegraphics[width=0.95\linewidth]{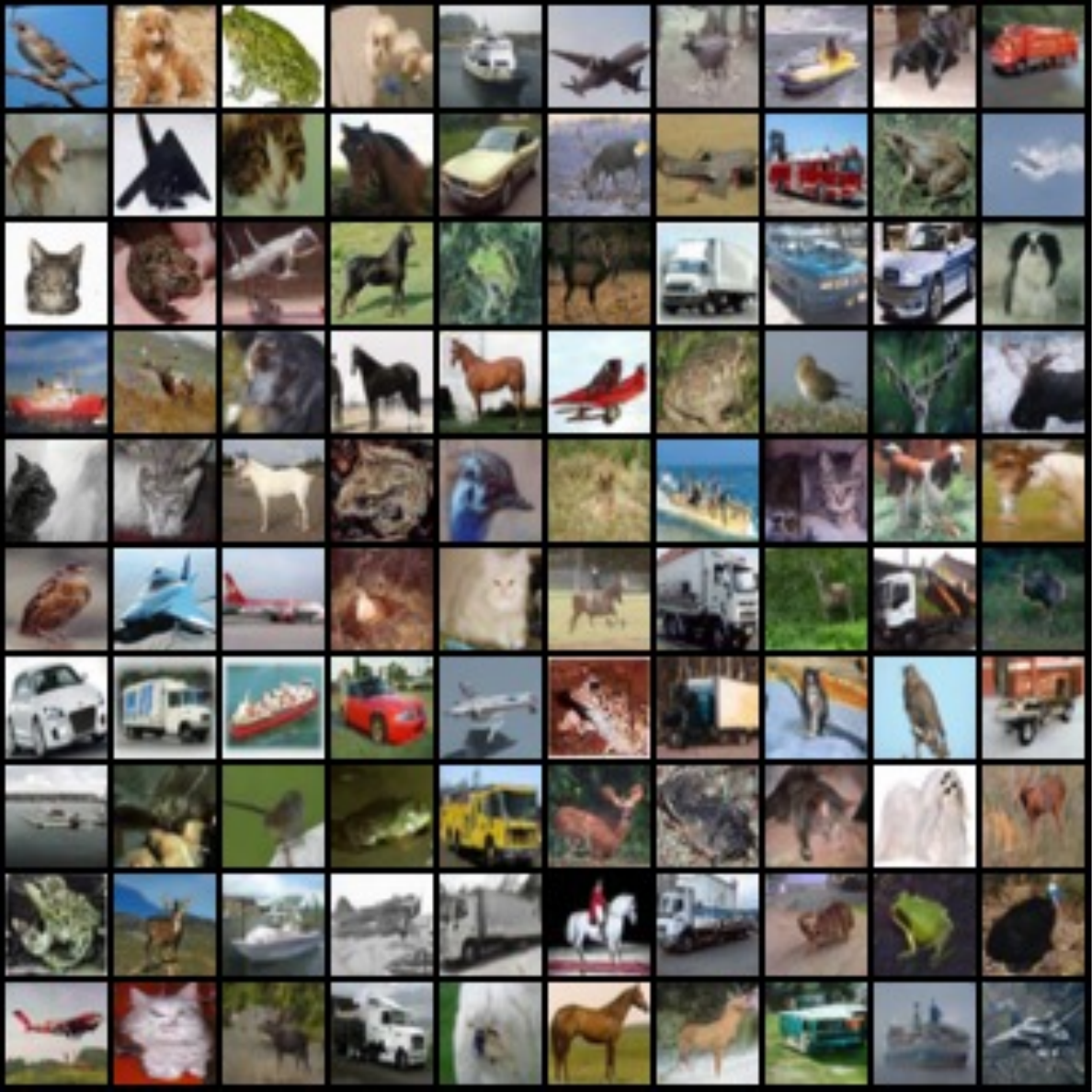}
\caption{ASCEND: randomly sampled results on CIFAR-10 (FID=2.98), 1000 sampling steps.}
\label{fig:ASCEND_cifar10}
\end{figure}

\begin{figure}[h]
\centering
\includegraphics[width=0.95\linewidth]{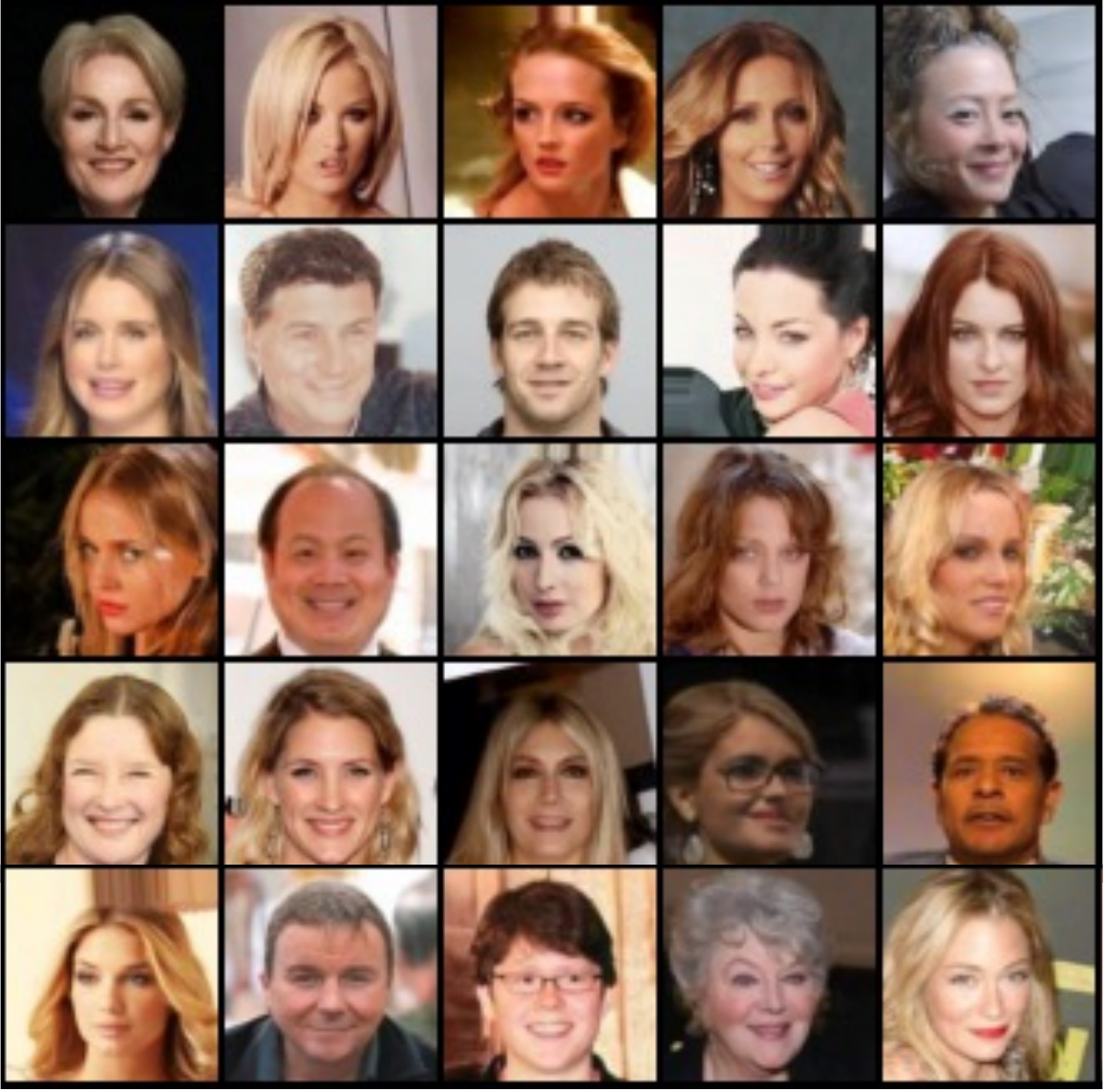}
\caption{ASCEND: randomly sampled results on CelebA$64\times64$ (FID=2.99), 1000 sampling steps.}
\label{fig:ASCEND_celeba64}
\end{figure}

\begin{figure}[ht]
\centering
\includegraphics[width=0.95\linewidth]{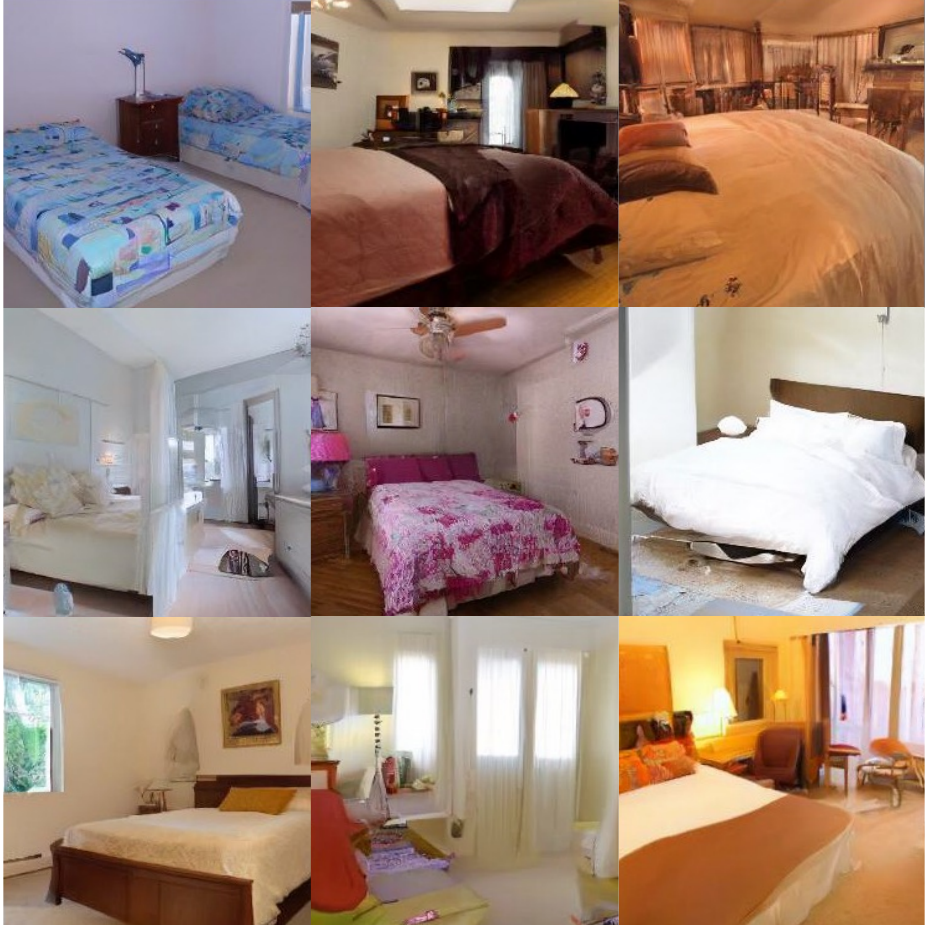}
\caption{ASCEND: randomly sampled results on LSUN Bedroom $256\times256$, 1000 sampling steps.}
\label{fig:ASCEND_bedroom256}
\end{figure}

\begin{figure}[]
\centering
\includegraphics[width=0.95\linewidth]{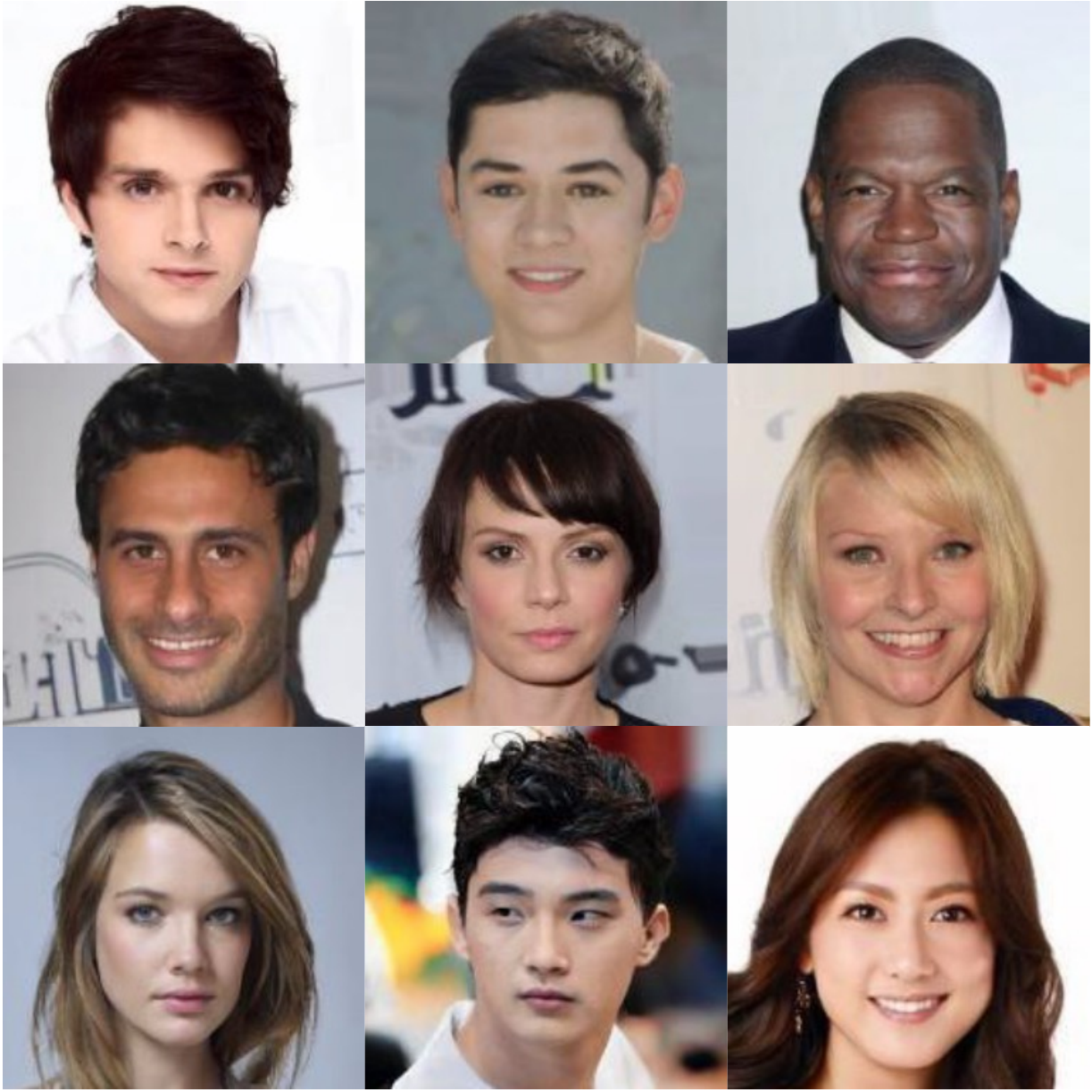}
\caption{ASCEND: randomly sampled results on CelebA $256\times256$, 1000 sampling steps.}
\label{fig:ASCEND_celeba256}
\end{figure}

\begin{figure*}[ht]
\centering
\includegraphics[width=0.95\linewidth]{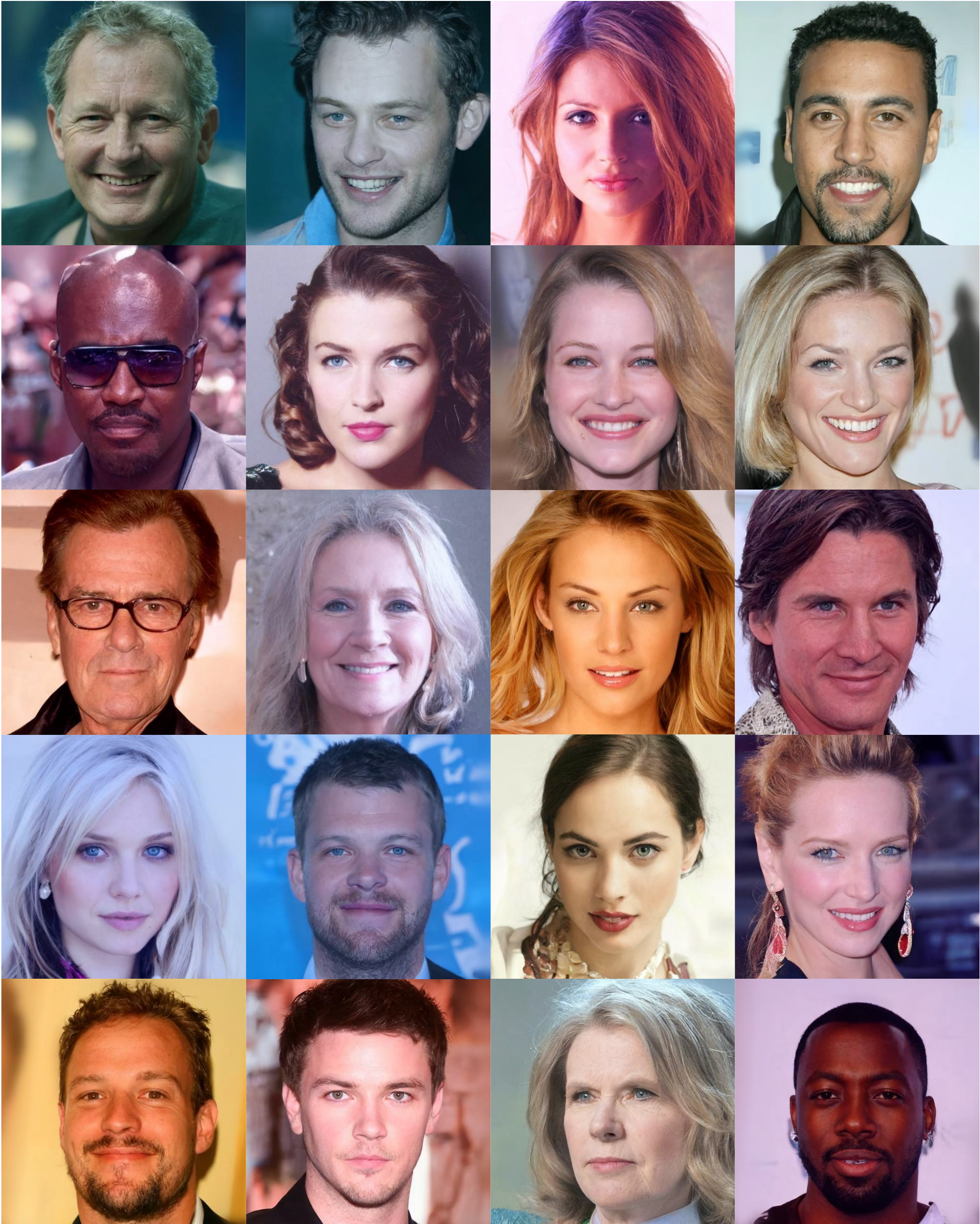}
\caption{ASCEND: randomly sampled results on CelebA-HQ $512\times512$, 1000 sampling steps.}
\label{fig:ASCEND_celebahq512}
\end{figure*}

\subsection{Additional ASCEND \textbf{$128\times128$ } Text-to-image Results}
\label{appendix_ascend_t2i}

\begin{figure*}[h]
    \centering
    \includegraphics[width=0.95\linewidth]{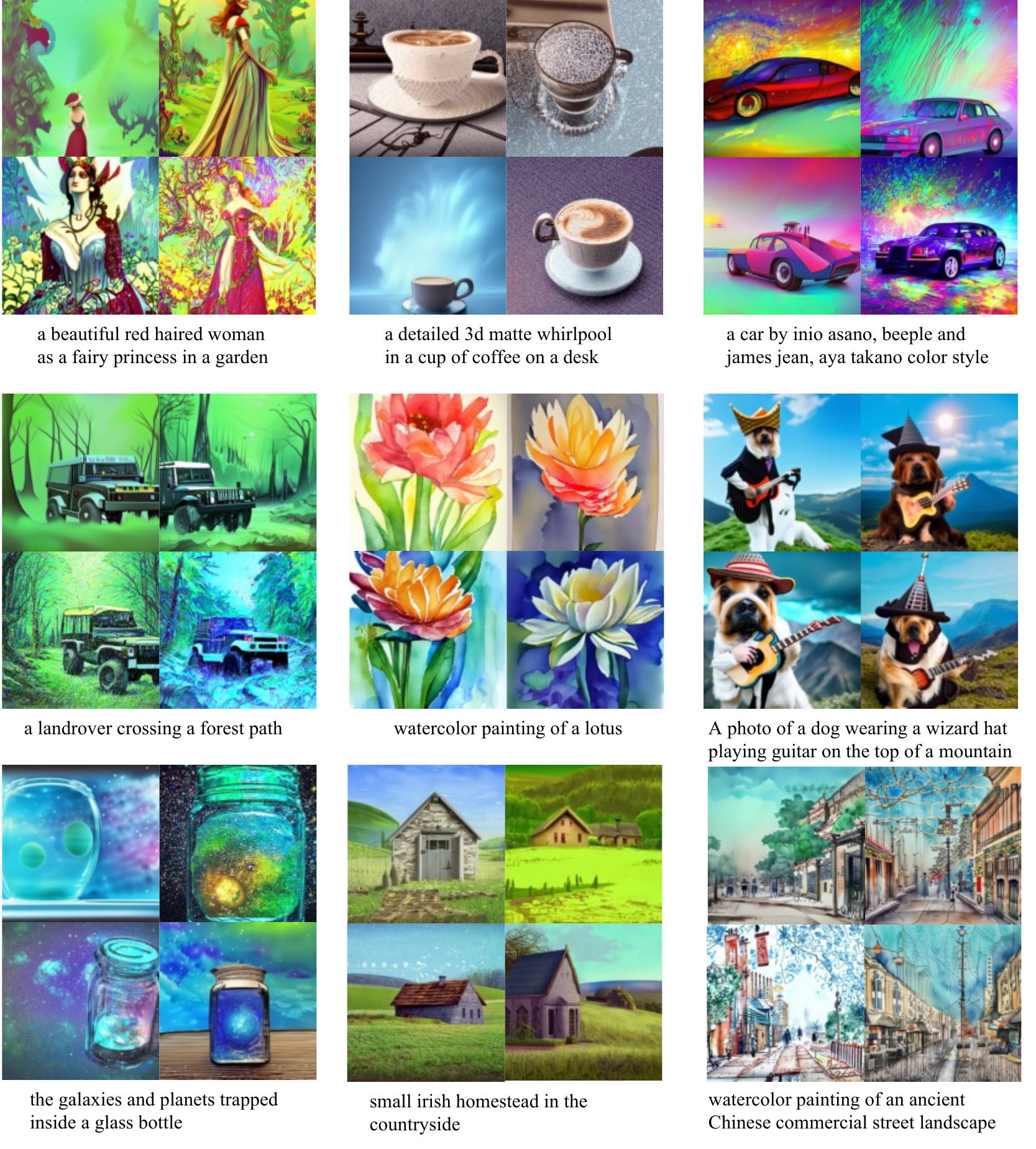}
    \caption{ASCEND $128\times128$ text-to-image samples for various text prompts. All images are sampled with 50 steps using DDIM.}
\label{fig:ASCEND_t2i128}
\end{figure*}

\end{document}